\documentclass[11pt,a4paper]{article}

\usepackage[utf8x]{inputenc}
\usepackage{ucs}
\usepackage{amsmath}
\usepackage{amsfonts}
\usepackage{amssymb}
\usepackage{graphicx}
\usepackage{caption}

\graphicspath{ {Figures/} }
\usepackage{tikz}
\usetikzlibrary{arrows,shapes,positioning,shadows,trees}

\usepackage{booktabs}
\usepackage{multirow}
\usepackage{natbib}

\newcommand{\dY}{\ensuremath{\mathcal{Y}}}
\newcommand{\Relev}{\ensuremath{\phi}}

\usepackage{authblk}
\usepackage{natbib}

\title{A Survey of Predictive Modelling under Imbalanced Distributions}

\author[1,2]{Paula Branco}
\author[1,2]{Lu\'{i}s Torgo}
\author[1,2]{Rita P. Ribeiro}

\affil[1]{LIAAD - INESC TEC} 
\affil[2]{ DCC - Faculdade de Ci\^{e}ncias - Universidade do Porto}
\affil[ ]{paobranco@gmail.com, ltorgo@dcc.fc.up.pt, rpribeiro@dcc.fc.up.pt}

\begin{document}
\maketitle

\begin{abstract}
Many real world data mining applications involve obtaining predictive models using data sets with strongly imbalanced distributions of the target variable. Frequently, the least common values of this target variable are associated with events that are highly relevant for end users (e.g. fraud detection, unusual returns on stock markets, anticipation of catastrophes, etc.). Moreover, the events may have different costs and benefits, which when  associated with the rarity of some of them on the available training data creates serious problems to predictive modelling techniques. This paper presents a survey of existing techniques for handling these important applications of predictive analytics. Although most of the existing work addresses classification tasks (nominal target variables), we also describe methods designed to handle similar problems within regression tasks (numeric target variables). In this survey we discuss the main challenges raised by imbalanced distributions, describe the main approaches to these problems, propose a taxonomy of these methods and refer to some related problems within predictive modelling.
\end{abstract}

\section{Introduction}

Predictive modelling is a data analysis task whose goal is to build a model of an unknown function $Y = f(X_1, X_2, \cdots, X_p)$, based on a training sample $\{\langle \mathbf{x}_i, y_i\rangle\}_{i=1}^n$ with examples of this function. Depending on the type of the variable $Y$, we face either a classification task (nominal $Y$) or a regression task (numeric $Y$). Models are obtained through an optimisation process that tries to find the "optimal" model parameters according to some criterion. The most frequent criteria are the error rate for classification and the mean squared error for regression. For some  real world applications it is of key importance that the obtained models are particularly accurate at some sub-range of the domain of the target variable. Examples include diagnostic of rare diseases, forecasting  rare extreme returns on financial markets, among many others. Frequently, these specific sub-ranges of the target variable are poorly represented on the available training sample. In these cases we face what is usually known as a problem of imbalanced data distributions, or imbalanced data sets. In other words, in these  domains the cases that are more important for the user are rare and few exist on the available training set. The conjugation of the specific preferences of the user with the poor representation of these situations creates problems to modelling approaches at several levels. Namely, we typically need  (i) special purpose evaluation metrics that are biased towards the performance of the models on these rare cases, and moreover, we need means for (ii) making the learning algorithms focus on these rare events. Without addressing these two questions, models will tend to be biased to the most frequent (and uninteresting for the user) cases, and the results of the "standard" evaluation metrics will not capture the competence of the models on these rare cases.

In this paper we provide a general definition for the problem of imbalanced domains that is suitable for both classification and regression tasks. We present an extensive survey of existing performance assessment measures and approaches to the problem of imbalanced data distributions.  Existing surveys address only the problem of imbalanced domains for classification tasks (e.g. \citet{kotsiantis2006handling, he2009learning, sun2009classification}). Therefore, the coverage of performance assessment measures and approaches to tackle both classification and regression tasks is an innovative aspect of our paper. Another key feature of our work is the proposal of a broader taxonomy of methods for handling imbalanced domains. Our proposal extends previous taxonomies by including post-processing strategies.

The main contributions of this work are: i) provide a general definition of the problem of imbalanced domains suitable for classification and regression tasks; ii) review the main performance assessment measures for classification and regression tasks under imbalanced domains; iii)  provide a taxonomy of existing approaches to tackle the problem of imbalanced domains both for classification and regression tasks; and iv) describe the most important techniques to address this problem.

The paper is organised as follows. Section~\ref{def} defines the problem of imbalanced data distributions and the type of existing approaches to address this problem.
Section~\ref{eval} describes several evaluation metrics that are biased towards performance assessment on the relevant cases in these domains.
Section~\ref{model} provides a taxonomy of the modelling approaches to imbalanced domains, describing some of the most important  techniques in each category.
Finally, Section \ref{relatedprob} explores some problems related with imbalanced domains and Section \ref{conc} concludes the paper.

\section{Problem Definition}
\label{def}


As we have mentioned before the problem of imbalanced data distributions occurs in the context of predictive tasks where the goal is to obtain a good approximation of the unknown function $Y = f(X_1, X_2, \cdots, X_p)$ that maps the values of a set of $p$ predictor variables into the values of a target variable. These approximations to the function are obtained using a training data set $D= \{\langle \mathbf{x}_i, y_i\rangle\}_{i=1}^n$. At the center of the problem of imbalanced distribution is the fact that the user assigns more importance to the performance of the obtained approximation on a subset of the range of values of the target variable $Y$. Let us express this user preference bias by an importance or relevance function $\phi()$ that maps the values of the target variable into a range of importance, where 1 is maximal importance and 0 minimum relevance,

\begin{equation}\label{fun:RelF}
\Relev(Y):\dY \to [0,1]
\end{equation}

\noindent where $\dY$ is the domain of the target variable $Y$.

Suppose the user defines a relevance threshold $t_R$ which sets the boundary above which the target variable values are relevant for the user. Let $D_R \in D$ be the subset of the training samples for which the relevance of the target value is high (or above $t_R$), i.e. $D_R = \{\langle \mathbf{x}_i, y_i\rangle\ \in D : \Relev(y_i) > t_R\}$, and $D_N \in D$ be the subset of the training sample with the normal (or less important) cases, i.e $D_N = \{\langle \mathbf{x}_i, y_i\rangle\ \in D : \Relev(y_i) \leq t_R\} = D\setminus D_R$.


The problem of imbalanced data sets can be described by the following assertions:

\begin{itemize}
\item $\phi(Y)$ is not uniform across the domain of $Y$
\item The cardinality of the set of examples $D_R$ is much smaller than the cardinality of $D_N$
\item Standard evaluation criteria for both learning the models and evaluating their performance assume an uniform $\Relev(Y)$, i.e. they are insensitive to $\Relev(Y)$.
\end{itemize}

In this context, we potentially have a situation where the obtained models are sub-optimal with respect to the user-preference biases, and moreover, the metrics used to evaluate them are not in accordance with these biases and thus may be misleading.


Regarding the evaluation issue, traditional metrics are not adequate as they do not take into account the user preferences. Several solutions have been proposed to address this problem and overcome existing difficulties, mainly for classification tasks.


With respect to the inadequacy of the obtained models a large number of solutions has also appeared in the literature. 
We propose a categorisation of these approaches that considers three types of strategies: (i) modifications on the learning algorithms, (ii) changes on the data before the the learning process takes place and finally (iii) transformations applied to the predictions of the learned models.

\section{Performance Metrics for Imbalanced Domains}
\label{eval}

Obtaining a model from data can be seen as a search problem guided by an evaluation criterion that establishes a preference ordering among different alternatives. The main problem of imbalanced data sets lies on the fact that they are often associated with an user preference bias towards the performance on cases that are poorly represented in the available data sample. Standard evaluation criteria tend to focus the evaluation of the models on the most frequent cases, which is against the user preferences on these tasks. In fact, the use of common metrics in imbalanced domains can lead to sub-optimal classification models \citep{he2009learning, W04, KM97} and might produce misleading conclusions since these measures are insensitive to skewed domains \citep{ranawana2006optimized, daskalaki2006evaluation}. As such, selecting proper evaluation metrics plays a key role in the task of correctly handling data imbalance. Adequate metrics should not only provide means to compare the models according to the user preferences, but can also be used to drive the learning of these models. 

As the problem of imbalanced domains has been addressed mainly in classification problems, there are far more solutions for this type of tasks. We start by addressing the problem of evaluation metrics in classification  
and then move to regression. 

Table~\ref{tab:MetricSummary} summarises the main references concerning performance assessment  proposals for imbalanced domains in classification and regression.

\begin{table}[!h]%
\resizebox{\textwidth}{!}{
\begin{tabular}{@{}cl@{}}
\toprule
\textbf{Task type (Section)}  & \multicolumn{1}{c}{\textbf{Main References}} \tabularnewline
\cmidrule{1-2}
\begin{tabular}{l}
\textbf{Classification}\\
(\ref{classmetrics})\end{tabular}
&
\begin{tabular}{@{}l@{}}
\citet{estabrooks2001mixture, kubat1998machine, bradley1997use}\\
\citet{PFK98, DG06}\\
\citet{garcia2008new, garcia2009index, garcia2010theoretical,ranawana2006optimized}\\
\citet{batuwita2009new, batuwita2012adjusted, hand2009measuring, thai2011new}
\end{tabular}\tabularnewline
\cmidrule{1-2}
\begin{tabular}{l}
 \textbf{Regression} \\
 (\ref{regresmetrics})
\end{tabular}
 &
\begin{tabular}{@{}l@{}}
\citet{zellner1986bayesian, cain1995real,christoffersen1997optimal}\\
\citet{crone2005utility, lee2008loss, HernandezOrallo20133395}\\
\citet{BB03, T05, TR07,TR09}\\
\citet{ Rib11}\\
\end{tabular}\tabularnewline

\bottomrule
\end{tabular}
}
\caption{Metrics for classification and regression, corresponding sections and main bibliographic references}
\label{tab:MetricSummary}
\end{table}

\subsection{ Metrics for Classification Tasks}
\label{classmetrics}

The confusion matrix for a two-class problem presents the results obtained by a given classifier (cf. Table~\ref{tab:ConfusionMatrix}). This table provides for each class the instances that were correctly classified, i.e. the number of True Positives (TP) and True Negatives (TN), and the instances that were wrongly classified, i.e. the number of False Positives (FP) and False Negatives (FN).

\begin{table}%
\begin{center}
\resizebox{0.4\textwidth}{!}{
\begin{tabular}{@{}cccc@{}}
\toprule
 & & \multicolumn{2}{c}{\textbf{Predicted}}  \tabularnewline
 \cmidrule{3-4}
& & Positive & Negative\tabularnewline
\cmidrule{2-4}
\multirow{2}{*}{\textbf{True}} & Positive & TP & FN\tabularnewline
 & Negative & FP & TN\tabularnewline
\bottomrule
\end{tabular}
}
\caption{Confusion matrix for a two-class problem.}
\label{tab:ConfusionMatrix}
\end{center}
\end{table}

\textit{Accuracy} (cf. Equation~\ref{eq:Acc}) and its complement \textit{error rate}  are the most frequently used metrics for estimating the performance of learning systems in classification problems. For  two-class problems, \textit{accuracy} can be defined as follows,

\begin{equation}\label{eq:Acc}
\begin{array}{rr}
accuracy = \frac{TP+TN}{TP+FN+TN+FP}
\end{array}
\end{equation}


Considering a user preference bias towards the minority (positive) class examples, \textit{accuracy} is not suitable because the impact of the least represented, but more important examples, is reduced when compared to that of the majority class. For instance, if we consider a problem where only 1\% of the examples belong to the minority class, an high \textit{accuracy} of 99\% is achievable by predicting the majority class for all examples. Yet, all minority class examples, the rare and more interesting cases for the user, are misclassified. This is worthless when the goal is the identification of the rare cases.


The metrics used in imbalanced domains must consider the user preferences and, thus, should take into account the data distribution. To fulfill this goal several performance measures were proposed. From Table~\ref{tab:ConfusionMatrix} the following measures {(cf. Equations~\ref{eq:TPrate}-\ref{eq:NPval})} can be obtained,

\begin{equation}\label{eq:TPrate}
\begin{array}{rrrrr}
true ~ positive ~ rate ~(recall~or~sensitivity): ~TP_{rate}=\frac{TP}{TP+FN}
\end{array}
\end{equation}

\begin{equation}\label{eq:TNrate}
\begin{array}{rrrrr}
true ~ negative ~ rate ~ (specificity~): ~ TN_{rate}=\frac{TN}{TN+FP}
\end{array}
\end{equation}

\begin{equation}\label{eq:FPrate}
\begin{array}{rrrrr}
false ~ positive ~ rate:~ FP_{rate}=\frac{FP}{TN+FP}
\end{array}
\end{equation}

\begin{equation}\label{eq:FNrate}
\begin{array}{rrrrr}
false~ negative ~ rate: ~ FN_{rate}=\frac{FN}{TP+FN}
\end{array}
\end{equation}

\begin{equation}\label{eq:PPval}
\begin{array}{rrrrr}
positive ~ predictive ~ value ~(precision~): ~ PP_{value}=\frac{TP}{TP+FP}
\end{array}
\end{equation}

\begin{equation}\label{eq:NPval}
\begin{array}{rrrrr}
negative ~ predictive ~value: ~ NP_{value}=\frac{TN}{TN+FN}
\end{array}
\end{equation}

However, as some of these measures exhibit a trade-off and it is impractical to simultaneously monitor several measures, new metrics have been developed, 
such as the \textit{F-measure} \citep{van1979information},
 the \textit{geometric mean} \citep{kubat1998machine} or the \textit{receiver operating characteristic} (\textit{ROC}) curve \citep{egan1975signal}.


The \textit{F-Measure} ($F_{\beta}$), a combination of both \textit{precision} and \textit{recall}, is defined as follows:
\begin{equation}\label{eq:FMeasure}
F_{\beta}=\frac{(1+\beta)^2 \cdot recall \cdot precision}{\beta ^2 \cdot recall + precision }
\end{equation}
\noindent where $\beta$ is a coefficient to adjust the relative importance of \textit{recall} with respect to \textit{precision} (if $\beta=1$ \textit{precision} and \textit{recall} have the same weight, large values of $\beta$ will increase the weight of \textit{recall} whilst values less than 1 will give more importance to \textit{precision}).


$F_{\beta}$ is commonly used and is more informative about the effectiveness of a classifier on predicting correctly the cases that matter to the user (e.g. \citet{estabrooks2001mixture}). This metric value is high when both \textit{recall} (a measure of completeness) and \textit{precision} (a measure of exactness) are high.




An also frequently used metric when dealing with imbalanced data sets is the \textit{geometric mean} (\textit{G-Mean}) which is defined as:

\begin{equation}\label{eq:GeomMean}
G-Mean=\sqrt{\frac{TP}{TP+FN} \times \frac{TN}{TN+FP}}=\sqrt{sensitivity \times specificity}
\end{equation}

\textit{G-Mean} is an interesting measure because it computes the \textit{geometric mean} of the accuracies of the two classes, attempting to maximise them while obtaining good balance.



\begin{figure}[!hbt]
\begin{center}
\resizebox{0.7\textwidth}{!}{
  \includegraphics{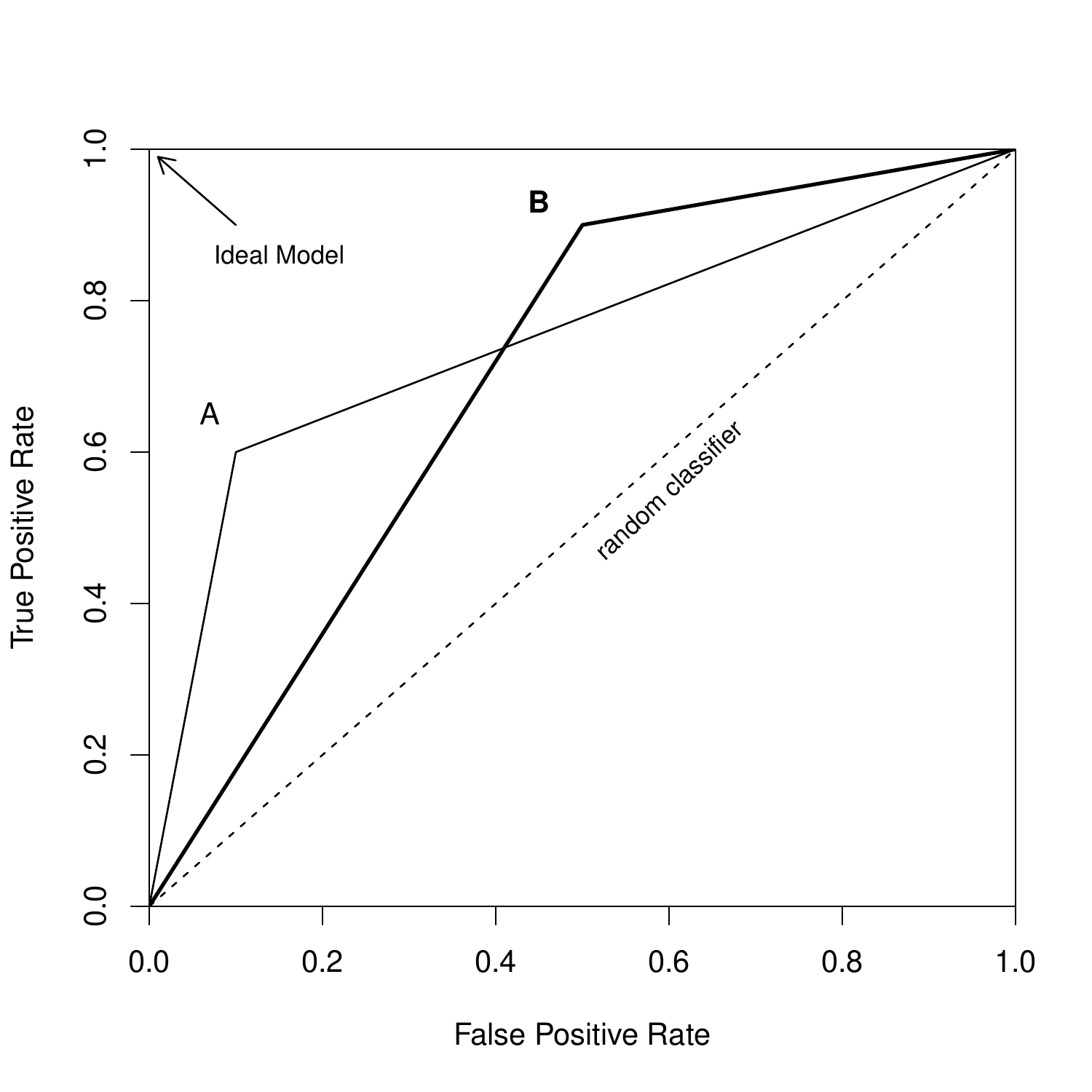}
}
\caption{\textit{ROC} curve of three classifiers: A, B and random.}
\label {fig:ROC}
\end{center}
\end{figure}

Two popular tools used in imbalanced domains are the \textit{receiver operating characteristics} (\textit{ROC}) curve (cf. Figure~\ref{fig:ROC}) and the corresponding area under the \textit{ROC} curve (\textit{AUC}) \citep{metz1978basic} . \citet{PFK98} proposed \textit{ROC} and \textit{AUC} as alternatives to \textit{accuracy}. The ROC curve allows the visualisation of the relative trade-off between 
benefits ($TP_{rate}$) and costs ($FP_{rate}$). The performance of a classifier for a certain distribution is represented by a single point in the \textit{ROC} space. A \textit{ROC} curve consists of several points each one corresponding to a different value of a decision/threshold parameter used for classifying an example as belonging to the positive class.

However, comparing several models through \textit{ROC} curves is not an easy task unless one of the curves dominates all the others~\citep{provost1997analysis}. Moreover, \textit{ROC} curves do not provide a single-value performance score which motivates the use of \textit{AUC}. The \textit{AUC} (cf. Equation \ref{eq:AUC}) allows the evaluation of the best model on average.  Still, it is not biased towards the minority class.

\begin{equation}\label{eq:AUC}
AUC=\frac{1+TP_{rate}-FP_{rate}}{2}= \frac{TP_{rate}+TN_{rate}}{2} 
\end{equation}

\textit{Precision-recall curves} (\textit{PR curves}) are recommended for highly skewed domains where \textit{ROC} curves may provide an excessively optimistic view of the performance~\citep{DG06}. \textit{PR curves} have the \textit{recall} and \textit{precision} rates represented on the axes. A strong relation between \textit{PR} and \textit{ROC} curves was found by~\citet{DG06}.



Several other measures were proposed for dealing with some particular disadvantages of the previously mentioned metrics. For instance, a metric called \textit{dominance} \citep{garcia2008new} (cf. Equation~\ref{eq:Dom}) was proposed to deal with the inability of \textit{AUC} and \textit{G-Mean} to explain how each class contributes to the overall performance.

\begin{equation}\label{eq:Dom}
dominance=TP_{rate}-TN_{rate}
\end{equation}

This measure ranges from $-1$ to $+1$.A value of $+1$ represents situations where perfect \textit{accuracy} is achieved on the minority (positive) class, but all cases of the majority class are missed. A value of $−1$ corresponds to the opposite situation. 

Another example is the \textit{index of balanced accuracy}
(\textit{IBA}) \citep{garcia2009index,
  garcia2010theoretical}
~(cf. Equation~\ref{eq:IBA}) which quantifies a trade-off between an index of how balanced both class accuracies are and a chosen unbiased measure of overall \textit{accuracy}.

\begin{equation}\label{eq:IBA}
IBA_{\alpha}(M) = (1 + \alpha \cdot dominance)M
\end{equation}

\noindent where $(1 + \alpha \cdot dominance)$ is the weighting factor and $M$ represents any performance metric. 

Several other metrics exist such as \textit{optimized precision}~\citep{ranawana2006optimized}, \textit{adjusted geometric mean}~\citep{batuwita2009new, batuwita2012adjusted}, \textit{H-measure}~\citep{hand2009measuring} or \textit{B42}~\citep{thai2011new}. All of them try to overcome some specific disadvantage detected in another metric when addressing
the challenge of assessing the performance in imbalanced domains.

\subsection{Metrics for Regression Tasks}
\label{regresmetrics}

Very few efforts have been made regarding evaluation metrics for regression tasks in imbalanced domains. Performance measures commonly used in regression, such as \textit{Mean Squared Error} (MSE) and \textit{Mean Absolute Deviation} (MAD) (cf. Equations~\ref{eq:MSE} and \ref{eq:MAD}) are not adequate to these specific problems. These measures assume an uniform relevance of the target variable domain and evaluate only the magnitude of the error.


\begin{equation}
\label{eq:MSE}
MSE=\frac{1}{n} \sum_{i=1}^{n} (y_{i}-\hat{y}_{i})^{2}
\end{equation}

\begin{equation}
\label{eq:MAD}
MAD=\frac{1}{n} \sum_{i=1}^{n} |y_{i}-\hat{y}_{i}|
\end{equation}






Although the magnitude of the numeric error is important, for tasks with imbalanced distribution of the target variable, the metric must also be sensitive to the errors location within the target variable domain, because as in classification tasks, users of these domains are frequently biased to the performance on poorly represented values of the target. A simple solution, such as the introduction of weights, would not fulfil this goal because it would neglect the errors of predicting a rare value when it is a normal one~\citep{Rib11}.

Within finance several attempts have been made for considering differentiated prediction costs through the proposal of asymmetric loss functions~\citep{zellner1986bayesian, cain1995real, christoffersen1996further, christoffersen1997optimal, crone2005utility, granger1999outline, lee2008loss}.
However, the proposed solutions, such as \textit{LIN-LIN} or \textit{QUAD-EXP} error metrics, all suffer from the same problem: they can only distinguish between over- and under-predictions. Therefore, they are still unsuitable for addressing the problem of imbalanced domains with a user preference bias towards some specific ranges of values.

\begin{figure}[!hbt]
\begin{center}
\resizebox{0.7\textwidth}{!}{
  \includegraphics{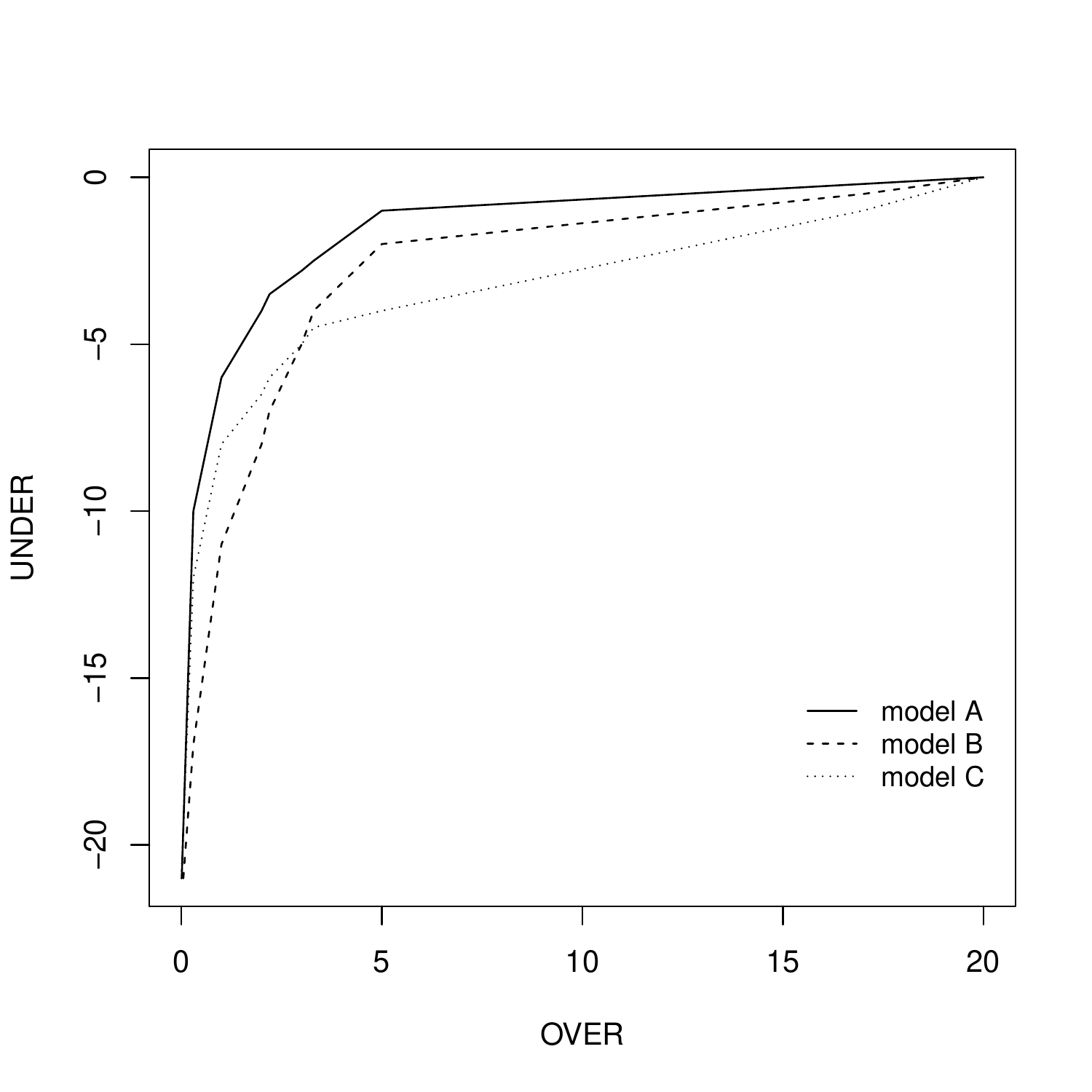}
}
\caption{\textit{RROC} curve of three models: A, B and C.}
\label {fig:RROC}
\end{center}
\end{figure}

Following the efforts made within classification, some attempts were made to adapt the existing notion of \textit{ROC} curves to regression tasks. One of these attempts is the \textit{ROC space for regression} (\textit{RROC} space)~\citep{HernandezOrallo20133395} which is motivated by the asymmetric loss often present on regression applications where both over-estimations and under-estimations entail different costs.
\textit{RROC} space is defined by plotting the total over-estimation and under-estimation on the $x$-axis and $y$-axis, respectively (cf. Figure \ref{fig:RROC}). \textit{RROC} curves are obtained when the notion of shift is used, which allows to adjust the model to an asymmetric operating condition by adding or subtracting a constant to the predictions.
The notion of dominance can also be assessed by plotting the curves of different regression models, similarly to \textit{ROC} curves in classification problems.
Other evaluation metrics were explored, such as the \textit{Area Over the} \textit{RROC curve} (\textit{AOC}) which was shown to be equivalent to the error variance. In spite of the importance of this approach, it still only distinguishes over  from under predictions.

\begin{figure}[!hbt]
\begin{center}
\resizebox{0.7\textwidth}{!}{
  \includegraphics{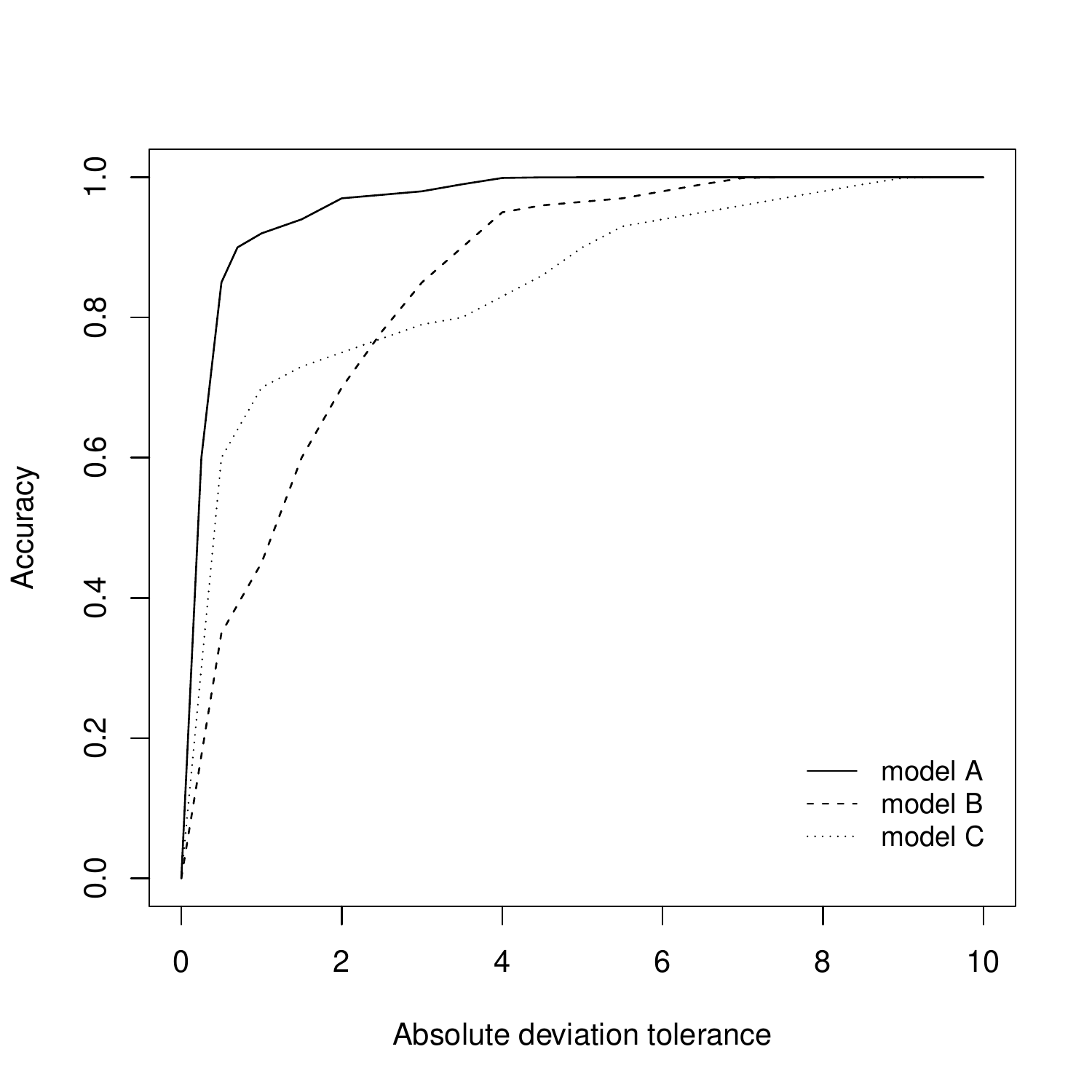}
}
\caption{\textit{REC} curve of three models: A, B and C.}
\label {fig:REC}
\end{center}
\end{figure}

Another relevant effort towards the adaptation of the concept of ROC curves to regression tasks was made by \citet{BB03} with the proposal of \textit{Regression Error Characteristic} (\textit{REC}) curves that provide a graphical representation of the cumulative distribution function (cdf) of the error of a model. 
These curves plot the error tolerance and the accuracy of a regression function which is defined as the percentage of points predicted within a given tolerance $\epsilon$. 
\textit{REC} curves illustrate the predictive performance of a model across the range of possible errors (cf. Figure~\ref{fig:REC}). 
The \textit{Area Over the Curve} (\textit{AOC}) can also be evaluated and is a biased estimate of the expected error of a model~\citep{BB03}.
\textit{REC} curves, although interesting, 
are still not sensitive to the error location across the target variable domain.

\begin{figure}[!hbt]
\begin{center}
\resizebox{0.6\textwidth}{!}{
  \includegraphics{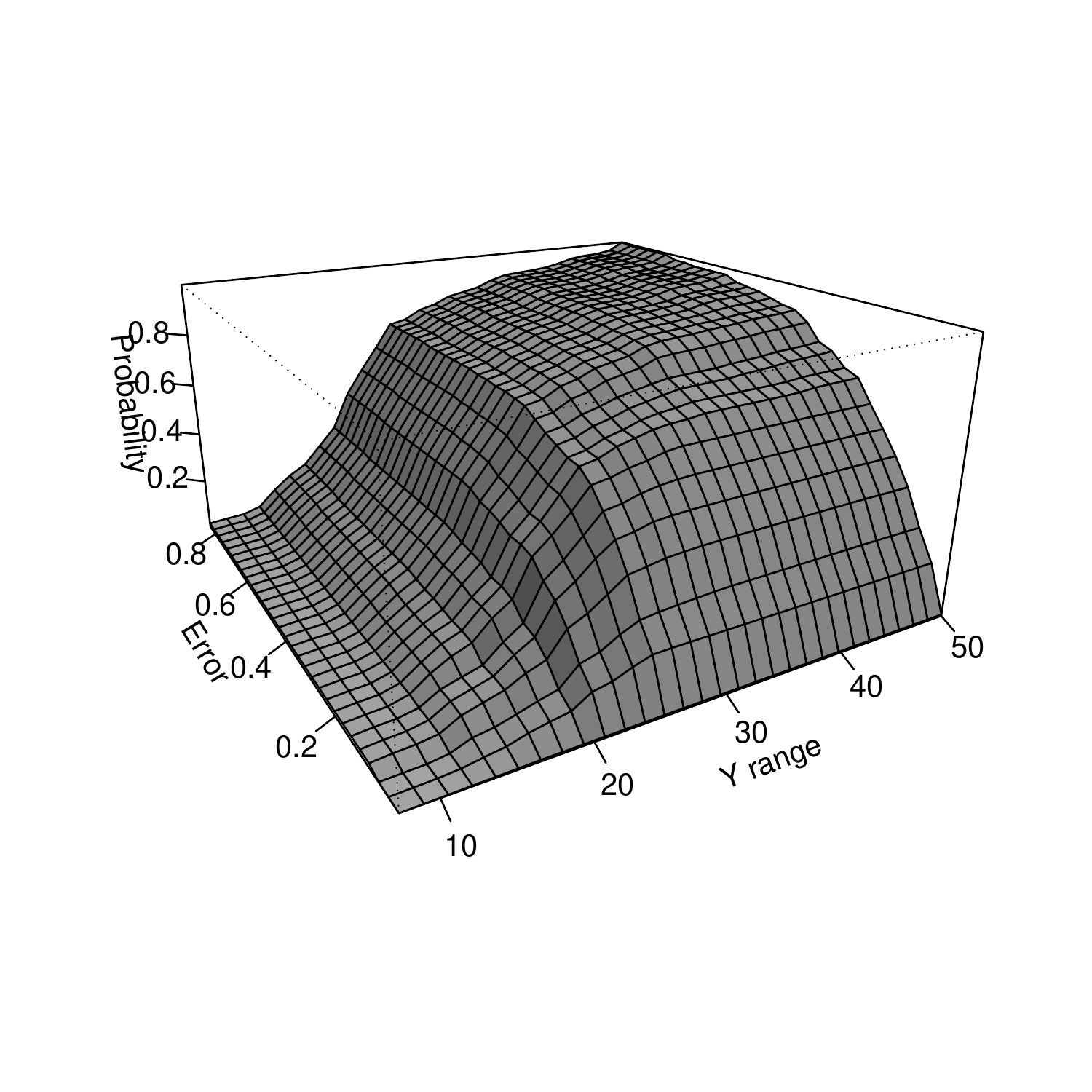}
}
\caption{An example of the \textit{REC} surface.}
\label {fig:RECS}
\end{center}
\end{figure}

To address this problem \textit{Regression Error Characteristic Surfaces} (\textit{RECS}) \citep{T05} were proposed. These surfaces incorporate an additional dimension into \textit{REC} curves representing the cumulative distribution of the target variable. 
\textit{RECS} show how the errors corresponding to a certain point of the REC curve are distributed across the range of the target variable (cf. Figure~\ref{fig:RECS}). This tool allows the study of the behaviour of alternative models for certain specific values of the target variable. By zooming on specific regions of REC surfaces we can carry out two types of analysis that are highly relevant for some application domains. The first involves checking how certain values of prediction error are distributed across the domain of the target variable, which tells us where this type of errors are more frequent. The second type of analysis involves inspecting the type of errors a model has on a certain range of the target variable that is of particular interest to us.

\begin{figure}
\centering
\begin{minipage}{.49\textwidth}
  \centering
  \includegraphics[width=\linewidth]{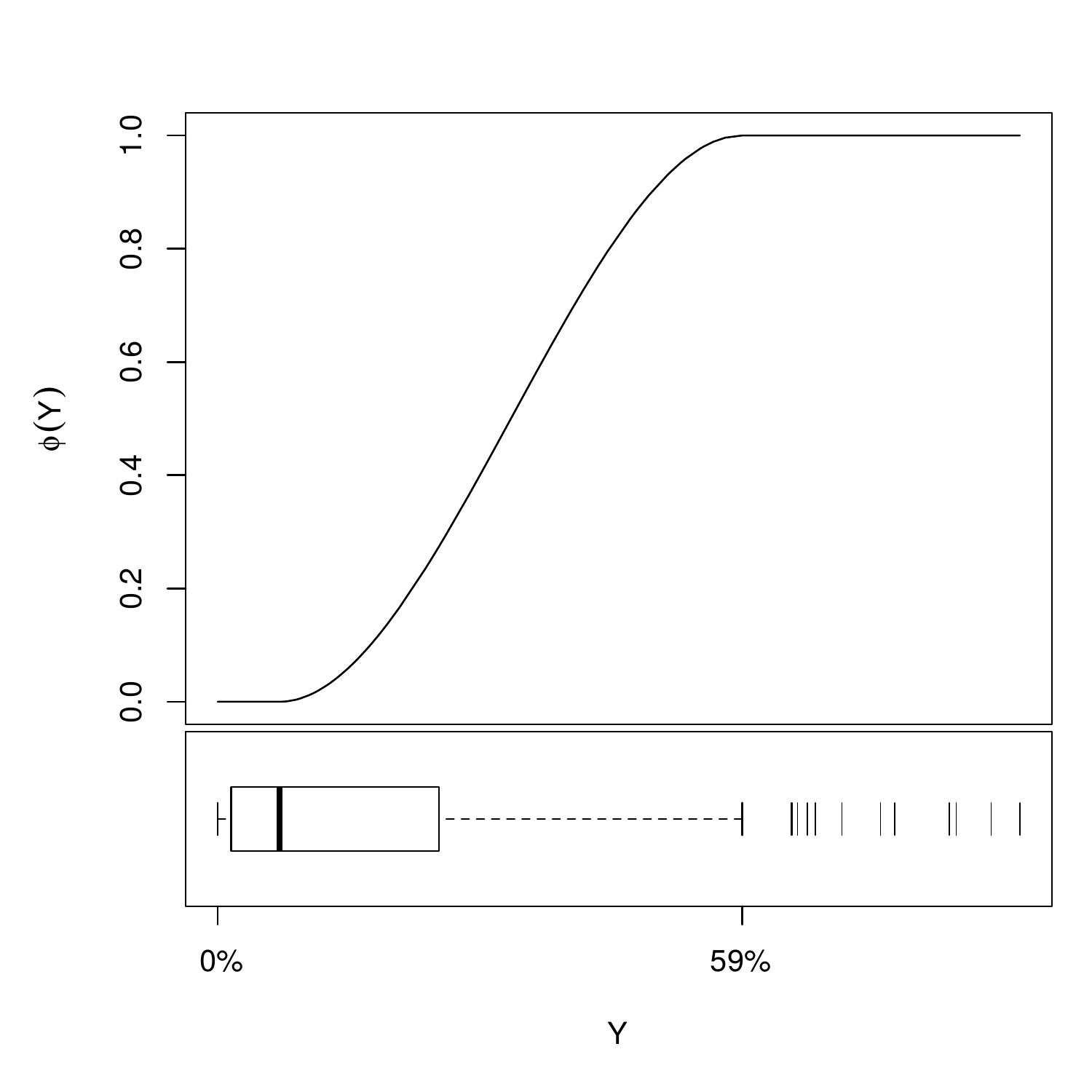}
  \caption{Relevance function $\phi$ automatically generated}
  \label{fig:a1Phi}
\end{minipage}%
\hspace{0.1cm}
\begin{minipage}{.49\textwidth}
  \centering
  \includegraphics[width=\linewidth]{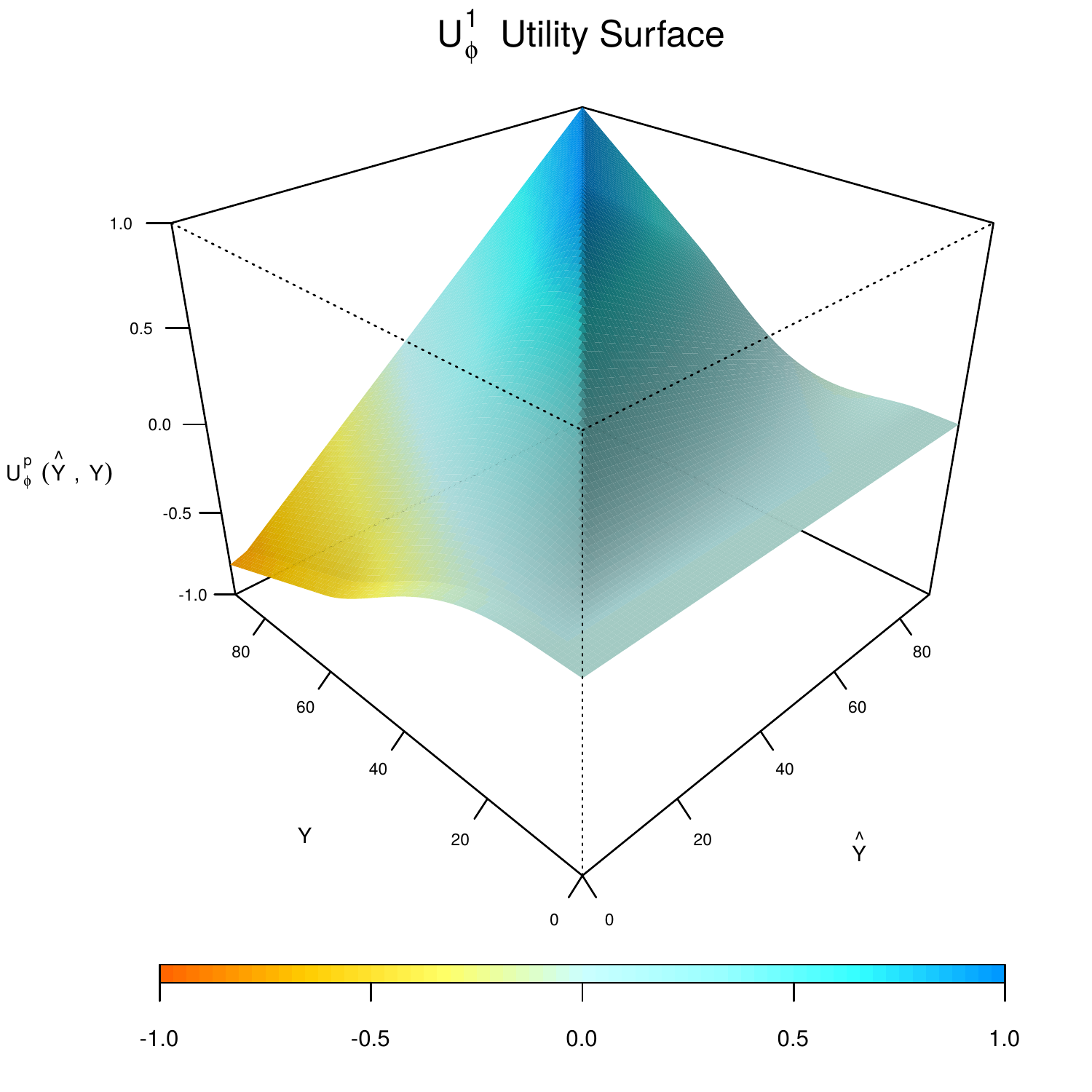}
  \caption{Utility surface obtained with relevance function $\phi()$ shown in Figure \ref{fig:a1Phi}}
  \label{fig:a1Surf}
\end{minipage}
\end{figure}

Another existing approach is the precision/recall evaluation framework, based on the concept of utility-based regression~\citep{Rib11,TR07}. 
Utility-based regression establishes the notion of relevance of the target variable values
and the existence of a non uniform relevance across the
domain of this variable. 
In this context, the
usefulness of a prediction dependes on both the numeric error of the
prediction (which is provided by a certain loss function $L(\hat{y},y)$) and the
relevance (importance) of the predicted $\hat{y}$ and true $y$
values. 
The relevance function, $\phi()$, is a continuous function as defined in Equation~\ref{fun:RelF} which expresses the importance of the target variable values.  Considering the goal of being accurate at
rare extreme values,
\citet{Rib11} 
describes some methods for automatically obtaining these
functions.
The methods are based on the
simple observation that, in these cases, the notion of relevance
is inversely proportional to the target variable probability.
Figure \ref{fig:a1Phi} shows an example of the relevance function $\phi$ in a data set where the high extreme values of the target variable are the most important, and Figure \ref{fig:a1Surf} shows the corresponding  utility surface .

Using this utility-based framework, the notions of precision and recall were adapted to regression problems with non-uniform relevance of the target values by~\citet{TR09} and~\citet{Rib11}. 
\citet{Rib11} defines the notion of event using the concept of utility. In this context, the ratios of the two metrics are also defined as functions of utility,   finally leading to  definitions of \textit{precision} and \textit{recall} for regression\footnote{Full details can be obtained in Chapter 4 of~\cite{Rib11}.}.
The notion of utility led to the proposal of other measures, such as the \textit{Mean Utility} and \textit{Normalized Mean Utility}~\citep{Rib11}. These metrics are derived from the utility and enable the comparison of different regression models according to the user preference bias. 

%
%



\section{Modelling Strategies for Handling Imbalanced Domains}
\label{model}




Imbalanced domains raise significant challenges when building predictive models. The scarce representation of the most important cases leads to models that tend to be more focused on the normal examples, neglecting the rare  events. Several strategies have been developed to address this problem, mainly in a classification setting.
We propose that the existing approaches to learn under imbalanced data distributions can be grouped into the following four  main categories:
\begin{itemize}
\item Data Pre-processing;
\item Special-purpose Learning Methods;
\item Prediction Post-processing;
\item Hybrid Methods.
\end{itemize}


Data Pre-processing approaches include solutions that pre-process the given imbalanced data set, changing the data distribution to make standard algorithms focus on the cases that are more relevant for the user.
These methods have the following advantages: (i) can be applied to any existing learning tool; and (ii) the chosen models are biased to the goals of the user (because the data distribution was previously changed to match these goals), and thus it is expected that the models are more interpretable in terms of these goals. The main inconvenient of this strategy  is that it may be difficult to relate the modifications in the data distribution with the target loss function.
This means that mapping the given data distribution into an optimal new distribution according to the user goals is not easy.

Special-purpose learning methods 
comprise solutions that change the existing algorithms to be able to learn from imbalanced data. The following are important advantages: (i) the user goals are incorporated directly into the models; and (ii) it is expected that the models obtained this way are more comprehensible to the user. The main disadvantages of these approaches are: (i) the user is restricted in his choice to the learning algorithms that have been modified to be able to optimise his goals, or has to develop new algorithms for the task; (ii) if the target loss function changes, the model must be relearned, and moreover, it may be necessary to introduce further modifications in the algorithm which may not be straightforward; and  (iii) it requires a deep knowledge of the learning algorithms implementations.

Prediction  Post-processing approaches use the original data set and a standard learning algorithm, only manipulating the predictions of the models according to the user preferences and the imbalance of the data.
As advantages, we can enumerate that:  (i) it is not necessary to be aware of the user preference biases at learning time; (ii) the obtained model can, in the future, be applied to different deployment scenarios (i.e. different loss functions), without the need of re-learning the models or even keeping the training data available; and  (iii) any standard learning tool can be used. However, these methods also have some drawbacks: (i) the models do not reflect the user preferences; (ii) the models 
interpretability is meaningless as they were obtained optimising a loss function that is not in accordance with the user preference bias.

\begin{figure}[!htbp]
\centering
\small
\resizebox{\textwidth}{!}{
\tikzset{
 basic/.style  = {draw, text width=4cm, drop shadow, font=\sffamily, rectangle},
  root/.style   = {basic, rounded corners=4pt, thin, align=center,
                   fill=gray!5},
  level 2/.style = {basic, rounded corners=8pt, thin,align=center, fill=orange!30,
                   text width=10em},
 level 3/.style = {basic, thin, rounded corners=2pt, align=left, fill=blue!60, text width=8.5em, fill=orange!20, node distance=1.5cm}
}

\begin{tikzpicture}[
  level 1/.style={sibling distance=40mm},
  edge from parent/.style={->,draw},
  >=latex]

\node[root] {\\Modelling Strategies for Imbalanced Domains\\\hfill\\}
  child {node[level 2] (c1) {Data \\Pre-processing}}
  child {node[level 2, fill=green!20] (c2) 
  {Special-purpose \\Learning Methods}
  }
  child {node[level 2, fill=blue!30] (c3) 
  {Prediction \\Post-processing}}
  child {node[level 2, fill=yellow!30] (c4) 
{Hybrid\\Methods}};

\begin{scope}[every node/.style={level 3}]
\node [below of = c1, xshift=5pt] (c11) {Re-sampling};
\node [below of = c11] (c12) {Active Learning};
\node [below of = c12] (c13) 
{Weighting the \\Data Space};


\node [below of = c3, xshift=5pt, fill=blue!15] (c31) {Threshold Method};
\node [below of = c31, fill=blue!15] (c32) 
{Cost-sensitive \\Post-processing};

\node [below of = c4, xshift=5pt, fill=yellow!20] (c41) 
{Re-sampling +\\ Special-purpose \\Learning Methods};

\end{scope}

\foreach \value in {1,2,3}
  \draw[->] (c1.190) |- (c1\value.west);


\foreach \value in {1,2}
  \draw[->] (c3.190) |- (c3\value.west);

\foreach \value in {1}
  \draw[->] (c4.190) |- (c4\value.west);

\end{tikzpicture}
}
\caption{Main modelling strategies for imbalanced domains.}
\label{fig:modellingTypes}
\end{figure}
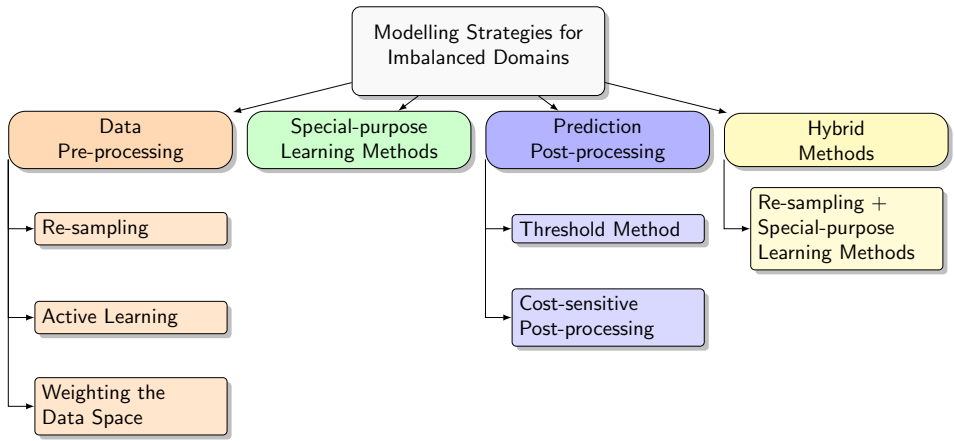

Approaches following these three types of strategies will be reviewed in Sections~\ref{dataLevel}, \ref{sec:algorithmLevel} and \ref{sec:predictionLevel},  and will include solutions for both classification and regression tasks. In Section~\ref{sec:Comb}  hybrid solutions will be addressed. Hybrid methods combine approaches of different types trying to take advantage of their best characteristics. Figure~\ref{fig:modellingTypes} synthesizes the different existing approaches within each of the categories.


\subsection{Data Pre-processing }
\label{dataLevel}



Pre-processing strategies consist of methods of using the available data set in a way that is more in accordance with the user preference biases. This means that instead of applying a learning algorithm directly to the provided training data, we will first somehow pre-process this data according to the goals of the user. Any standard learning algorithm can  be applied to the pre-processed data set. 

Existing data pre-processing approaches can be grouped into three main  types:
\begin{itemize}
\item \textbf{re-sampling:} change the data distribution of the data set forcing the learner to focus on the least represented examples;
\item \textbf{active learning:} actively selecting the best (more valuable) samples to learn, leaving the ones with less information to improve the learner performance;
\item \textbf{weighting the data space:} modify the training set distribution using information concerning misclassification costs, such that the learned model avoids costly errors. 
\end{itemize}


Table~\ref{tab:preSummary} summarizes the main bibliographic references for data pre-processing strategies.

\begin{table}%
\resizebox{\textwidth}{!}{
\begin{tabular}{@{}cll@{}}
\toprule
\multicolumn{2}{c}{\textbf{Strategy type (Section)}} & \multicolumn{1}{c}{\textbf{Main References}} \tabularnewline
\cmidrule{1-3}

\multirow{16}{*}{\begin{tabular}{l}
 \textbf{Re-sampling} \\
 (\ref{sec:resampling})
\end{tabular}} & 
  Random Under/Over-sampling 
& \begin{tabular}{@{}l@{}} \citet{CBOK02, drummond2003c4}\\ \citet{estabrooks2004multiple,seiffert2010rusboost}; \\ \citet{chen2004using, wang2009diversity}; \\ \citet{chang2003statistical, tao2006asymmetric}; \\ \citet{torgo2013smote}\end{tabular} \tabularnewline
\cmidrule{2-3}

 & Distance Based&  \citet{Chyi2003class, mani2003knn}\tabularnewline
 \cmidrule{2-3}
 
 &  Data Cleaning Based  & \begin{tabular}{@{}l@{}}\citet{KM97, laurikkala2001improving}; \\ \citet{batista2004study, naganjaneyulu2013novel}\end{tabular}\tabularnewline
 \cmidrule{2-3}
 &   Recognition Based  &\begin{tabular}{@{}l@{}} \citet{chawla2004editorial,zhuang2006parameter}; \\ 
 \citet{raskutti2004extreme};\\ 
 \citet{ japkowicz2000learning,bellinger2012one};\\
 \citet{ lee2006novelty,zhuang2006parameterestim}\end{tabular} \tabularnewline
 \cmidrule{2-3}
 
 & Cluster Based  & \begin{tabular}{@{}l@{}} \citet{jo2004class, yen2006under, yen2009cluster};\\ \citet{cohen2006learning}\end{tabular} \tabularnewline
 \cmidrule{2-3}
 
 &    Synthesising New Data
& \begin{tabular}{@{}l@{}}\citet{lee1999regularization, lee2000noisy, CBOK02, liu2007generative}; \\ \citet{menardi2010training,chawla2003smoteboost};\\
 \citet{martinez2012sneom,wang2009diversity};\\ \citet{torgo2013smote}\end{tabular}\tabularnewline
 \cmidrule{2-3}
 
 &   Adaptive Synthetic Sampling
  &  \begin{tabular}{@{}l@{}}\citet{batista2004study, verbiest2012improving}; \\ \citet{hu2009msmote, zhang2011novel}; \\ \citet{ barua2012mwmote, ramentol2012smote, ramentol2012smoteRSB}; \\ \citet{bunkhumpornpat2012dbsmote, nakamura2013lvq};\\ \citet{bunkhumpornpat2009safe,  han2005borderline}; \\ \citet{he2008adasyn,maciejewski2011local} \end{tabular}\tabularnewline
 \cmidrule{2-3}
 
 &  Evolutionary Sampling  
 & \begin{tabular}{@{}l@{}}\citet{garcia2006proposal, doucette2008gp};   \\ \citet{garcia2009evolutionary, drown2009evolutionary};\\ \citet{del2004multistrategy,  yong2012research};\\ \citet{maheshwari2011new, derrac2012evolutionary}; \\ \citet{ galar2013eusboost}\end{tabular}\tabularnewline
 \cmidrule{2-3}
 
 &   Re-sampling Combinations  & \begin{tabular}{@{}l@{}} \citet{stefanowski2008selective, napierala2010learning}; \\ \citet{ songwattanasirismoute}; \\ \citet{yang2012active,li2008hybrid};\\ \citet{vasu2011hybrid, bunkhumpornpat2011mute};\\ \citet{jeatrakul2010classification,liu2006boosting}; \\ \citet{ mease2007cost,chen2010ramoboost}\end{tabular}\tabularnewline
 \cmidrule{1-3}
 
\multicolumn{2}{l}{\begin{tabular}{l}
 \textbf{Active Learning}  \\
 (\ref{sec:activeLearn})
\end{tabular} }  & \begin{tabular}{@{}l@{}}\citet{ertekin2007active, ertekin2007learning, zhu2007active}\\ \citet{ertekin2013virtual, mi2013imbalanced} \end{tabular}\tabularnewline
 \cmidrule{1-3}
\multicolumn{2}{l}{\begin{tabular}{l}
\textbf{ Weighting the Data Space} \\
 (\ref{sec:weightDataSpace})
\end{tabular} }  &\citet{zadrozny2003cost, wang2010boosting} \tabularnewline
\bottomrule
\end{tabular}
}
\caption{Pre-processing strategy types, corresponding sections and main bibliographic references}
\label{tab:preSummary}
\end{table}



\subsubsection{Re-sampling}
\label{sec:resampling}



Applying re-sampling strategies to obtain a more balanced data distribution is an effective solution to the imbalance problem \citep{estabrooks2004multiple, batuwita2010efficient, fernandez2008study,fernandez20102}.

However, changing the data distribution may not be as easy as expected. 
Decide what is the optimal distribution is not straightforward as it is a domain dependent decision. 
Moreover, it was proved for classification tasks that a perfectly balanced distribution does not always provide optimal results~\citep{weiss2003learning}. In this context, some solutions were proposed to find the right amount of re-sampling for a data set~\citep{weiss2003learning, chawla2005wrapper, chawla2008automatically}.

For classification problems, changing the class distribution of the training data improves classifiers performance on an imbalanced context because it imposes non-uniform misclassification costs. This equivalence between the two concepts of altering the data distribution and the misclassification cost ratio is well-known and was first pointed out by~\citet{breiman1984classification}.


The existing re-sampling strategies  are based on a diverse set of techniques such as: random under/over-sampling, distance methods, data cleaning approaches, clustering algorithms, synthesising new data or evolutionary algorithms. We now briefly describe the most significant re-sampling strategies.


Two of the most simple re-sampling approaches that can be applied are under- and over-sampling. The first one removes data from the original data set reducing the sample size, while the second one adds data increasing the sample size. In random under-sampling, a random set of majority class examples are discarded. This may eliminate useful examples leading to a worse performance. Oppositely, in random over-sampling, a random set of copies of minority class examples  is added to the data. This may increase the likelihood of overfitting, specially for higher over-sampling rates~\citep{CBOK02, drummond2003c4}. Moreover, it may decrease the classifier performance and increase the computational effort.




Random under-sampling was also used in the context of ensembles. Namely, it was combined with
 boosting \citep{seiffert2010rusboost}, bagging \citep{wang2009diversity, chang2003statistical, tao2006asymmetric} and was applied to both classes in random forests in a method named Balanced Random Forest (BRF) \citep{chen2004using}. 

For regression tasks, \citet{torgo2013smote} perform random under-sampling of the common values as a strategy for addressing the imbalance problem. This method uses a relevance function and an user defined threshold to determine which are the common and uninteresting values that should be under-sampled.

Despite the potential of randomly selecting examples, under- and over-sampling strategies can also be carried out by other, more informed, methods.
For instance, under-sampling can be accomplished resorting to distance evaluations~\citep{Chyi2003class, mani2003knn}. These approaches perform under-sampling based on a certain distance criteria that determines which are the examples from the majority class to include in the training set. 
These strategies are very time consuming which is a major disadvantage, specially when dealing with large data sets.

Under-sampling can also be achieved through data cleaning methods. The main goal of these methods is to identify possibly noisy examples or overlapping regions and then decide on the removal of examples. One of those methods  uses Tomek links~\citep{tomek1976two} which consist of points that are each other's closest neighbours, but do not share the same class label. This method allows for two options: only remove Tomek links examples belonging to the majority class or eliminate Tomek links examples of both classes~\citep{batista2004study}. The notion of Condensed Nearest Neighbour Rule (CNN) \citep{cnn} was also applied to perform under-sampling~\citep{KM97}. CNN is used to find a subset of examples consistent with the training set, i.e., a subset that correctly classifies the training examples using a 1-nearest neighbour classifier. CNN and Tomek links methods were combined in this order by \citet{KM97} in a strategy called One-Sided-Selection (OSS), and in the reverse order in a proposal of \citet{batista2004study}. 




Recognition-based methods as one-class learning or autoencoders offer the possibility to perform the most extreme type of under-sampling where all the examples from the majority class are removed. In this type of approach, and contrary to discrimination-based inductive learning, the model is learned using only examples of the target class, and no counter examples are included. This lack of examples from the other class(es) is the key distinguishing feature between recognition-based and discrimination-based learning. 


One-class learning tries to set up boundaries which surround the target concept. This method starts by measuring the similarity between the target class and an object. Classification is then performed using a threshold on the obtained similarity score.
One-class learning methods have the disadvantage of requiring the tuning of the threshold imposed on
the similarity. In fact, this is a sensitive issue because if
 we choose a too narrow threshold the minority class examples
are disregarded. However, too wide thresholds may lead to including examples from the majority class. Therefore, establishing an efficient threshold is vital with this method. Also, some learners actually need examples from more than one class and are unable to adapt to this method. Despite all these possible disadvantages, recognition-based learning algorithms have been proved to provide good prediction performance in most domains.
Developments made in this context include one-class SVMs (e.g. \citet{scholkopf2001estimating, manevitz2002one, raskutti2004extreme, zhuang2006parameter, zhuang2006parameterestim, lee2006novelty}) and the use of an autoencoder (or autoassociator) (e.g. \citet{japkowicz1995novelty, japkowicz2000learning}). 

\citet{bellinger2012one} investigated the performance variations of binary and one-class classifiers for different levels of imbalance. The results on both artificial and real world data sets showed that as the level of imbalance increased, the performance of binary classifiers decreased, whereas the performance of one-class classifiers stayed relatively stable. 


Imbalanced domains can influence the performance and the efficiency of clustering algorithms~\citep{xuan2013cluster}. However, 
due to their flexibility,
several approaches appeared for dealing with imbalanced data sets using clustering methods . For instance, the cluster-based oversampling (CBO) algorithm proposed by \citet{jo2004class} addresses both the imbalance problem and the problem of small disjuncts. Small disjuncts are subclusters of a certain class which have a low coverage, i.e., classify only few examples~\citep{holte1989concept}.
CBO consists of clustering the training data of each class separately with the k-means technique and then performing random over-sampling in each cluster. All majority class clusters are over-sampled until they reach the  cardinality of the largest cluster of this class. Then the minority class clusters are over-sampled until both classes are balanced maintaining all minority class subclusters with the same number of examples. Several other proposals based on clustering techniques exist (e.g. \citet{yen2006under, yen2009cluster, cohen2006learning}).

Another important approach for dealing with the imbalance problem as a pre-processing step, is the generation of new synthetic data. Several methods exist for building new synthetic examples. Most of the proposals are focused on classification tasks. Synthesising new data has several known advantages~\citep{CBOK02, menardi2010training}, namely: (i) reduces the risk of overfitting which is introduced when replicas of the examples are inserted in the training set; (ii) improves the ability of generalisation which was compromised by the over-sampling methods. The methods for synthesising new data can be organized in two groups: (i) one that uses interpolation of existing examples, and (ii) another that introduces perturbations.

A famous method that uses interpolation is the synthetic minority over-sampling technique - SMOTE~\citep{CBOK02}. 
SMOTE algorithm over-samples the minority class by generating new synthetic data. This technique is then combined with a certain percentage of random under-sampling of the majority class that depends on a user defined parameter. Artificial data is created using an interpolation strategy that introduces a new example along the line segment joining a seed example and one of its $k$ minority class nearest neighbours. The number of minority class neighbours ($k$) is another user defined parameter. For each minority class example a certain number of examples is generated according to a predefined over-sampling percentage. 
%

SMOTE algorithm has been applied with several different classifiers and was also integrated with boosting \citep{chawla2003smoteboost} and bagging \citep{wang2009diversity}.


SMOTE  generates synthetic examples with the positive class label disregarding the
negative class examples which may lead to overgeneralization~\citep{yen2006under, maciejewski2011local, yen2009cluster}.
This strategy may be specially problematic in the case of highly skewed class distributions where the minority class examples are very sparse, thus resulting in a greater chance of class mixture. 

The group of techniques that introduces perturbations for generating new data does not suffer from this problem. \citet{lee1999regularization} proposed an over-sampling method that produces noisy replicates of the rare cases while keeping the majority class unchanged. The synthetic examples are generated by adding normally distributed noise to the minority class examples.
This simple strategy was tested with success, and a new version was developed by \citet{lee2000noisy}. This new approach generates, for a given data set, multiple versions of training sets with added noise. Then, an average of multiple model estimates is obtained.

Another framework, named ROSE (Random Over Sampling Examples), for dealing with the problem of imbalanced classification was presented by \citet{menardi2010training} based on a smoothed bootstrap re-sampling technique. 
ROSE generates a more balanced and completely new data set from the given training set combining over- and under-sampling. One observation is draw from the training set by giving the same probability to both existing classes. A new example is generated in the neighbourhood of this observation,
using a width for the neighbourhood determined by a chosen smoothing matrix.



Several other proposals exist for classification tasks (e.g. \citet{liu2007generative, martinez2012sneom}). However, for regression problems only one method for generating new synthetic data was proposed. \citet{torgo2013smote} have adapted the SMOTE algorithm to regression tasks. Three key components of the SMOTE algorithm required adaptation for regression: (i) how to define which are the relevant observations and the "normal" cases; (ii) how to generate the new synthetic examples (i.e. over-sampling); and (iii) how to determine the value of the target variable in the synthetic examples.
Regarding the first issue, a relevance function and a user-specified threshold were used to define $D_R$ and $D_N$ sets. The observations in $D_R$ are over-sampled, while cases in $D_N$ are under-sampled. For the generation of new synthetic examples  the same interpolation method used in SMOTE for classification was applied. Finally, the target value of each synthetic example was calculated as an weighted average of the target variable values of the two seed examples. The weights were calculated as an inverse function of the distance of the generated case to each of the two seed examples.

Some drawbacks  identified in the SMOTE algorithm 
motivated the appearance of several variants of this method~\citep{barua2012mwmote, han2005borderline, bunkhumpornpat2009safe, chawla2003smoteboost, he2008adasyn, maciejewski2011local, ramentol2012smote, verbiest2012improving,stefanowski2007improving}.

We can identify three main types of SMOTE variants: (i) application of some pre- or post- processing before or after the use of SMOTE; (ii) apply SMOTE only in some selected regions of the input space; or (iii) introducing small modifications to the SMOTE algorithm. 
Most of the first type of SMOTE variants start by applying the SMOTE algorithm and, afterwards, use a post-processing mechanism for removing some data. Examples of this type of approaches include: SMOTE+Tomek~\citep{batista2004study}, SMOTE+ENN~\citep{batista2004study}, SMOTE+FRST~\citep{ramentol2012smote} or SMOTE+RSB~\citep{ramentol2012smoteRSB}. An exception is the Fuzzy Rough Imbalanced Prototype Selection (FRIPS)~\citep{verbiest2012improving} method that  pre-processes the data set before applying the SMOTE algorithm.
The second type of SMOTE variants only generates synthetic examples in specific regions that are considered useful for the learning algorithms. As the notion of what is a good region is not straightforward, several strategies were developed. Some of these variants focus the synthesising effort on the borders between classes while others try to find which are the harder to learn instances and concentrate on these ones. Examples of these approaches  are: Borderline-SMOTE~\citep{han2005borderline}, ADASYN~\citep{he2008adasyn}, Modified Synthetic Minority Oversampling Technique (MSMOTE)~\citep{hu2009msmote}, MWMOTE~\citep{barua2012mwmote}, FSMOTE~\citep{zhang2011novel}, among others.
Regarding the last type of SMOTE variants, 
some modifications are introduced in the way SMOTE generates the synthetic examples. For instance, the synthetic examples may be generated closer or further apart from a seed depending on some measure. The following proposals are examples within this group:  Safe-Level-SMOTE~\citep{bunkhumpornpat2009safe}, Safe Level Graph~\citep{bunkhumpornpat2013safe}, LN-SMOTE~\citep{maciejewski2011local} and DBSMOTE \citep{bunkhumpornpat2012dbsmote}.

Another approach to re-sampling concerns the use of Evolutionary Algorithms (EA). These algorithms started to be applied to imbalanced domains as a strategy to perform under-sampling through a prototype selection (PS) procedure (e.g. \citet{garcia2006proposal, garcia2009evolutionary}).

\citet{garcia2006proposal} made one of the first contributions with a new evolutionary method proposed for balancing the data set. The method presented uses a new fitness function designed to perform a prototype selection process. Some proposals have also emerged in the area of heuristics and metrics for improving several genetic programming classifiers performance in imbalanced domains \citep{doucette2008gp}.

However, EA have been used for more than under-sampling. More recently, Genetic Algorithms (GA) and clustering techniques were combined to perform both under and over-sampling~\citep{maheshwari2011new, yong2012research}. Evolutionary under-sampling has also been combined with boosting \citep{galar2013eusboost}.

Finally, several other interesting methods have appeared which combine some of the previous techniques~\citep{stefanowski2008selective, bunkhumpornpat2011mute,songwattanasirismoute,yang2012active}. For instance, \citet{jeatrakul2010classification} presents a method that uses Complementary Neural Networks (CMTNN) to perform under-sampling and combines it with SMOTE. The combination of strategies was also applied to ensembles (e.g. \citet{liu2006boosting,mease2007cost,chen2010ramoboost}).

Some attention has also been given to SVMs, leading to proposals such as the one of \citet{kang2006eus} where an ensemble of under-sampled SVMs is presented. Multiple different training sets are built by sampling examples
from the majority class and combining them with the minority class examples. Each training set is used for training an individual SVM classifier. The ensemble is produced by aggregating the outputs of all individual classifiers. Another similar approach is the EnSVM \citep{liu2006boosting} which adopts a rebalance strategy combining the over-sampling strategy of SMOTE algorithm  and under-sampling to form a number of new training sets while using all the positive examples.
Then, an ensemble of SVMs is built. 

Several ensembles have been adapted and combined with re-sampling approaches to better tackle the problem of imbalanced domains. Essentially, for every type of ensembles, some attempt has been made. 
For a more complete review on ensembles for the class imbalance problem see \citet{galar2012review}.

\subsubsection{Active Learning}
\label{sec:activeLearn}



Active learning is a semi-supervised  strategy in which the learning algorithm is able to interactively obtain information from the user. Although this method is traditionally used with unlabelled data, it can also be applied when all class labels are known. In this case, the active learning strategy provides the ability of actively selecting the best, i.e. the most informative, examples to learn from.  

Several approaches for imbalanced domains based on active learning
have been proposed~\citep{ertekin2007active, ertekin2007learning,
  zhu2007active, ertekin2013virtual}. These approaches are
concentrated on SVM learning systems and are based on the fact that,
for this type of learners, the most informative examples are the ones
closest to the hyperplane. 
 
This property is used to guide under-sampling by selecting the most informative examples , i.e., choosing the examples closer to the hyperplane.




More recent developments try to combine active learning with other techniques 
to further improve learners performance.
\citet{ertekin2013virtual} presents a novel adaptive over-sampling
algorithm named Virtual Instances Resampling Technique Using Active
Learning (VIRTUAL), that combines the benefits of over-sampling and
active learning. Contrary to traditional re-sampling methods, which are
applied before the training stage, VIRTUAL generates synthetic
examples for the minority class during the training
process. Therefore,
the need for a separate pre-processing step is discarded. In the context of learning with SVMs, VIRTUAL outperforms competitive over-sampling techniques both in terms of generalisation performance and computational complexity. \citet{mi2013imbalanced} developed a method that combines SMOTE and active learning with SVMs.


Some efforts have also been made for integrating active learning with other classifiers. \citet{hu2012active} proposed an active learning method for imbalance data using the Localized Generalization Error Model (L-GEM) of radial basis function neural networks (RBFNN).

\subsubsection{Weighting the Data 
Space}
\label{sec:weightDataSpace}

The strategy of weighting the data space is a way of implementing cost-sensitive learning. In fact, misclassification costs are applied to the given data set with the goal of selecting the best training distribution. Essentially, this method is based on the fact that changing the original sampling distribution  by multiplying each case by a factor that is proportional to its importance (relative cost), allows any standard learner to accomplish expected cost minimisation on the original distribution.
Although it is a simple technique and easy to apply, it also has  some drawbacks. There is a risk of model overfitting and it is also possible that the real cost values are unavailable which can introduce the extra difficulty of exploring effective cost setups. 

This approach has a strong theoretical foundation, building on the \textit{Translation Theorem} derived by \citet{zadrozny2003cost}. Namely, to obtain a modified distribution biased towards the costly classes, the training set distribution is modified with regards to misclassification costs.
\citet{zadrozny2003cost} presented two different ways of accomplishing this conversion: in a transparent box or in a black box way. In the first, the weights are provided
to the classifier while for the second a careful subsampling is performed according to the same weights. The first approach cannot be applied to an arbitrary learner, while the second one results in severe overfitting if re-sampling with replacement is used. Thus, to overcome the drawbacks of the later approach,  the authors have presented a method called \textit{cost-proportionate rejection sampling} which accepts each example in the input sample with probability proportional to its associated weight.

\citet{wang2010boosting} proposes an ensemble of SVMs with asymmetric misclassification costs. The proposed system works by modifying the base classifier (SVM) using costs and uses boosting as the combination scheme.

\subsection{Special-purpose Learning Methods}
\label{sec:algorithmLevel}

The approaches at this level consist of solutions that modify existing algorithms to provide a better fit to the imbalanced training data. The task of developing a solution based on algorithm modifications is not an easy one. It requires a deep knowledge of both the learning algorithm and the target domain. In order to perform a modification on a selected algorithm, it is essential to understand why it fails when the distribution is skewed. Also, some of the adaptations assume that a cost/cost-benefit matrix is known for different error types, which is frequently not the case. On the other hand, these methods have the advantage of being very effective in the contexts for which they were designed. 

Existing solutions for dealing with imbalanced domains at the learning level are focused on the introduction of modifications in the algorithm preference criteria.

Table~\ref{tab:algoSummary} summarizes the main bibliographic references for strategies involving modifications of algorithms.

\begin{table}%
\resizebox{\textwidth}{!}{
\begin{tabular}{@{}ccl@{}}
\toprule
\multicolumn{2}{c}{\textbf{Strategy type (Section)}} & \multicolumn{1}{c}{\textbf{Main References}} \tabularnewline
 \cmidrule{1-3}
\multicolumn{2}{l}{\begin{tabular}{l}
 \textbf{Special-purpose Learning Methods} \\
 (\ref{sec:algorithmLevel})
\end{tabular}}  &\begin{tabular}{@{}l@{}}\citet{maloof2003learning, akbani2004applying, tang2009svms};\\ \citet{weiguo2012improved, zhou2006training};\\ \citet{oh2011error, castro2013novel};\\ \citet{sun2007cost, song2009improved, chen2004using};\\ \citet{joshi2001evaluating, hwang2011new};\\ 
\citet{ alejo2007improving, cao2013pso};\\
\citet{wu2003class, imam2006z}; \\ \citet{tang2006granular, batuwita2010fsvm} \\ \citet{li2009improved, barandela2003strategies};\\ \citet{ huang2004learning, liu2010robust, tan2003multi}; \\\citet{ cieslak2012hellinger,rodriguez2012disturbing};\\ \citet{wu2005kba, xiao2012dynamic};\\ \citet{ cieslak2008learning,Rib11};\\
\citet{torgo2003predicting, ribeiro2003predicting}\end{tabular}\tabularnewline
\bottomrule
\end{tabular}
}
\caption{Special-purpose Learning Methods, corresponding section and main bibliographic references}
\label{tab:algoSummary}
\end{table}


The incorporation of benefits and/or costs (negative benefits) in existing algorithms, as a way to express the utility of different predictions, is one of the known approaches to cope with imbalanced domains. This includes the well known cost-sensitive algorithms for classification tasks which directly incorporate costs in the learning process. In this case, the goal of the prediction task is to minimise the total cost, knowing that misclassified examples may have different costs.  In an imbalanced context, the cost of misclassifying a minority class example is superior to the cost of misclassifying a majority class example and, usually, there is no cost associated with making a correct prediction.


The research literature includes several works describing the adaptation of different classifiers in order to make them cost-sensitive.  
For decision trees, the impact of the incorporation of costs under imbalanced domains was addressed by \citet{maloof2003learning}
Regarding support vector machines several ways of integrating costs have been considered such as assigning different penalties to false negatives and positives \citep{akbani2004applying} or including a weighted attribute strategy \citep{yuanhong2009cost} among others \citep{weiguo2012improved}.
Regarding neural networks, the possibility of making them cost-sensitive has also been considered (e.g. \citet{zhou2006training, alejo2007improving, oh2011error}).
A Cost-Sensitive Multilayer Perceptron (CSMLP) algorithm was proposed by \citet{castro2013novel} for asymmetrical learning of MLPs via a modified (backpropagation) weight update rule. \citet{cao2013pso} present a framework for improving the performance of cost-sensitive neural networks that uses Particle Swarm Optimization (PSO) for optimizing misclassification cost, feature subset and intrinsic structure parameters. \citet{alejo2007improving} propose two strategies for dealing with imbalanced domains using RBF neural networks which include a cost function in the training phase.

Ensembles have also been considered in the cost-sensitive framework to handle imbalanced domains. Several ensemble methods have been successfully adapted to include costs during the learning phase. However, boosting was the most extensively explored. 
AdaBoost is the most representative algorithm of the boosting family. When the class distribution is imbalanced, AdaBoost biases the learning (through the weights) towards the majority class, as it contributes more to the overall accuracy. Several proposals appeared which modify AdaBoost weight update process by incorporating cost items so that examples from different classes are treated unequally. Important proposals in the context of imbalanced distributions are:  RareBoost~\citep{joshi2001evaluating}, AdaC1, AdaC2 and AdaC3~\citep{sun2007cost}, and BABoost~\citep{song2009improved}. All of them modify the AdaBoost algorithm by introducing costs in the used weight updating formula. These proposals differ in how they modify the  update rule.
Random Forests have also been adapted to better cope with imbalanced domains undergoing a cost-sensitive transformation. \citet{chen2004using} proposes a method called Weighted Random Forest (WRF) for dealing with highly-skewed class distributions based on the Random Forest algorithm. WRF strategy operates by assigning a higher misclassification cost to the minority class. For an extensive review on ensembles for handling class imbalance  see \citet{galar2012review}.






Several other solutions exist that also modify the preference criteria of the algorithms while not relying directly on the definition of a cost/cost-benefit matrix.
Regarding SVMs, several proposals try to bias the algorithm so that the hyperplane is further away from the positive class because the skew associated with imbalanced data sets pushes the hyperplane closer to the positive class. \citet{wu2003class} accomplish this  with an algorithm that changes the kernel function. 
Fuzzy Support Vector Machines for Class Imbalance Learning (FSVM-CIL) was a method proposed by \citet{batuwita2010fsvm}. This algorithm is based on an SVM variant for handling the problem of outliers and noise called FSVM \citep{lin2002fuzzy}
and improves it for also dealing with imbalanced data sets.
Potential Support Vector Machine (P-SVM) differs from standard SVM learners by defining a new objective function and constraints. 
An improved P-SVM algorithm~\citep{li2009improved} was proposed to better cope with imbalanced data sets. 




\textit{k}-NN learners were also adapted to  cope with the imbalance problem. \citet{barandela2003strategies} present a weighted distance function to be used in the classification phase of \textit{k}-NN without changing the class distribution. This method assigns different weights to the respective classes and not to the individual prototypes. Since more weight is given to the majority class, the distance to minority class examples becomes much lower than the distance to examples from the majority class. This biases the learner to find their nearest neighbour among examples of the minority class.


A new decision tree algorithm - Class Confidence Proportion Decision Tree (CCPDT) - was proposed by \citet{liu2010robust}. CCPDT is robust and insensitive to class distribution and generates rules that are statistically significant. The algorithm adopts a new proposed measure,
 called Class Confidence Proportion (CCP), which forms the basis of CCPDT. CCP measure is embedded in the information gain and used as the splitting criteria.  In this algorithm, a new approach , using Fisher exact test, to prune branches of the tree that are not statistically significant is presented.

Hellinger distance was introduced as a decision tree splitting criterion to build Hellinger Distance Decision Trees (HDDT) \citep{cieslak2008learning}. This proposal was shown to be insensitive towards class distribution skewness.
More recently, \citet{cieslak2012hellinger} recommended the use of bagged HDDTs as the preferred method for dealing with imbalanced data sets when using decision trees. 

For regression tasks, some works have addressed the problem of imbalanced domains by changing the splitting criteria of regression trees (e.g. \citet{torgo2003predicting, ribeiro2003predicting}). 




In \citet{wu2005kba} the Kernel Boundary Alignment algorithm (KBA) is proposed. This method adjusts the boundary towards the majority class by modifying the kernel matrix generated by a kernel function according to the imbalanced data distribution.


An ensemble method for learning over multi-class imbalanced data sets, named ensemble Knowledge for Imbalance Sample Sets (eKISS), was proposed by \citet{tan2003multi}. This algorithm was specifically designed to increase classifiers sensitivity without losing the corresponding specificity. The eKISS approach combines the rules of the base classifiers to generate new classifiers for final decision making. 

Recently, more sophisticated approaches were proposed as the Dynamic Classifier Ensemble method for Imbalanced Data (DCEID) presented by \citet{xiao2012dynamic}. DCEID combines dynamic ensemble learning with cost-sensitive learning and is able to adaptively select the more appropriate ensemble approach.
%

For regression problems one work exists that is able to tackle the problem of imbalanced domains through an utility-based algorithm. 
The utility-based Rules (ubaRules) approach was proposed by \citet{Rib11}. ubaRules is an utility-based regression rule ensemble system designed for obtaining models biased according to a specific utility-based metric. The system main goal is to obtain accurate and interpretable predictions in the context of regression problems with non-uniform utility. It consists in two main steps: generation of different regression trees, which are converted to rule ensembles, and selection of the best rules to include in the final ensemble. An utility function is used as criterion at several stages of the algorithm.

All these algorithm modification strategies are specifically designed to address the problem of imbalanced domains and have great potential. However, some disadvantages exist, such as: i) an often unavailable cost/cost-benefit matrix; ii) the need of a deep knowledge of the selected learner to accomplish a good modification of the preference criteria and iii) the difficulty of using an already existing method with a different learning system which contrasts with pre-processing approaches.

\subsection{Prediction Post-processing}
\label{sec:predictionLevel}

For dealing with imbalanced domains at the post-processing level, we will consider two main types of solutions:
\begin{itemize}
\item \textbf{threshold method:} 
uses the ranking provided by a score, that expresses the degree to which an example is a member of a class, to produce several learners by varying the threshold for class membership;
\item \textbf{cost-sensitive post-processing:} associates costs to prediction errors and minimizes the expected cost.
\end{itemize}

Table~\ref{tab:posSummary} summarizes the main bibliographic references of post-processing strategies.

\begin{table}%
\begin{center}
\resizebox{0.6\textwidth}{!}{
\begin{tabular}{@{}cll@{}}
\toprule
\multicolumn{2}{c}{\textbf{Strategy type (Section)}}  & \multicolumn{1}{c}{\textbf{Main References}} \tabularnewline
\cmidrule{1-3}
\multicolumn{2}{l}{\begin{tabular}{l}
 \textbf{Threshold Method}  \\
(\ref{sec:thresh})
\end{tabular}}  &\citet{maloof2003learning, W04}\tabularnewline
\cmidrule{1-3}
\multicolumn{2}{l}{\begin{tabular}{l}
 \textbf{Cost-sensitive Post-processing} \\
 (\ref{sec:costSensPost})
\end{tabular}}  &\begin{tabular}{@{}l@{}}  \citet{hernandez2012soft, orallo2014prob}\end{tabular}\tabularnewline
\bottomrule
\end{tabular}
}
\caption{Post-processing strategy types, corresponding sections and main bibliographic references}
\label{tab:posSummary}
\end{center}
\end{table}


\subsubsection{Threshold Method}
\label{sec:thresh}

Some classifiers are named soft classifiers because they provide a score which expresses the degree to which an example is a member of a class. This score can, in fact, be used as a threshold to generate other classifiers. This task can be accomplished by varying the threshold for an example belonging to a class \cite{W04}. A study of this method \citep{maloof2003learning} concluded that the operations of moving the decision threshold, applying a sampling strategy, and adjusting the cost matrix produce classifiers with the same performance.

\subsubsection{Cost-sensitive Post-processing}
\label{sec:costSensPost}


Several methods exist for making models cost-sensitive in a post hoc manner. This type of strategy was mainly explored for classification tasks and aims at changing only the model predictions for making it cost-sensitive (e.g. \citet{Dom99, sinha2004evaluating}). This means that these approaches could potentially be applicable to imbalanced data distributions. However, to the best of our knowledge,  these methods have never been applied or evaluated on these tasks.

In regression , introducing costs at a post-processing level has only recently been proposed \citep{bansal2008tuning, zhao2011extended}. It is an issue still under-explored with few limited solutions. Similarly to what happens in classification, no progress was yet made for evaluating these solutions in imbalanced domains.
However,  one interesting proposal called reframing~\citep{hernandez2012soft, orallo2014prob} was recently presented. Although not developed specifically for imbalanced domains, this framework aims at adjusting the predictions of a previously built model to different data distributions. Therefore, it is also potentially suitable for being applied to the problem of imbalanced domains.
The notion of reframing was established  as the process of applying a previously built model to a new operating context by the proper transformation of inputs, outputs and patterns. The reframing framework acts at a post-processing level, changing the obtained predictions  by adapting them to a different distribution.

The reframing method essentially consists of two steps:
\begin{itemize}
\item the conversion of any traditional crisp regression model with one parameter into a soft regression model with two parameters, seen as a normal conditional density estimator (NCDE), by the use of enrichment methods; 
\item the reframing of an enriched soft regression model to new contexts by an instance-dependent optimisation of the expected loss derived from the conditional normal distribution.
\end{itemize}
\subsection{Hybrid Methods}
\label{sec:Comb}

In recent years, several methods involving the combination of some of the basic approaches described in the previous sections, have appeared in the research literature. Due to their characteristics these methods can be seen as hybrid methods to handle imbalanced distributions. They try to capitalise on some of the main advantages of the different approaches we have described previously.   


Existing hybrid approaches combine the use of re-sampling strategies with special-purpose learning algorithms. 
Table~\ref{tab:hybridSummary} summarizes the main bibliographic references concerning these strategies.


\begin{table}%
\resizebox{\textwidth}{!}{
\begin{tabular}{@{}cll@{}}
\toprule
\multicolumn{2}{c}{\textbf{Strategy type (Section)}} &  \multicolumn{1}{c}{\textbf{Main References}} \tabularnewline
\cmidrule{1-3}
\multicolumn{2}{l}{\begin{tabular}{l}
 \textbf{Re-sampling and Special-purpose Learning Methods} \\
 (\ref{sec:resamplingAlgorithm})
\end{tabular}}  &\begin{tabular}{@{}l@{}}\citet{phua2004minority,kotsiantis2003mixture}; \\ \citet{estabrooks2001mixture, estabrooks2004multiple};\\  \citet{yoon2005unsupervised, liu2009exploratory}\end{tabular}\tabularnewline

\bottomrule
\end{tabular}
}
\caption{Hybrid strategies, corresponding sections and main bibliographic references}
\label{tab:hybridSummary}
\end{table}


\subsubsection{Re-sampling and Special-purpose Learning Methods}
\label{sec:resamplingAlgorithm}

%


One of the first  hybrid strategies was presented by \citet{estabrooks2001mixture} and \citet{estabrooks2004multiple}. The motivation for this proposal is related to the fact that a perfectly balanced data may not be optimal and that the right amount of over/under-sample to apply is difficult to determine. To overcome these difficulties, a mixture-of-experts framework was proposed \citep{estabrooks2001mixture, estabrooks2004multiple} in an architecture with three levels: a classifier level, an expert level and an output level. The system has two experts in the expert level: an under-sampling expert and an over-sampling expert. The architecture incorporates 10 classifiers on the over-sampling expert and another 10 classifiers on the under-sampling expert. All these classifiers are trained in data sets re-sampled at different rates of over and under-sampling, respectively. 
At the classifier level an elimination strategy is applied for removing the learners that are considered unreliable according to a predefined test. Then a combination scheme is applied both at the expert and output levels. These combination schemes use the following simple heuristic: if one of the classifiers decides that the example is positive so does the expert, and if one of the two experts decides that the example is positive so does the output level. This strategy is clearly heavily biased towards the minority (positive) class. 

A different idea involving re-sampling and the combination of different learners was proposed by~\citet{kotsiantis2003mixture}. The proposed approach uses a facilitator agent and three learning agents each one with its own learning system. The facilitator starts by filtering the features of the data set. The filtered data is then passed to the three learning agents. Each learning agent re-samples the data set, learns using the respective system (Naive Bayes, C4.5 and 5NN) and returns the predictions for each instance back to the facilitator agent. Finally, the facilitator makes the final prediction according to majority voting.

In the proposal of \citet{phua2004minority} re-sampling is performed and then stacking and boosting are used together. The applied re-sampling strategy  partitions the data set into eleven new data sets which include all the minority class examples and a portion of the majority class examples. The proposed system uses three different learners (Naive Bayes, C4.5 and back-propagation classifier) each one processing the eleven partitions of the data. Bagging is used to combine the classifiers trained by the same algorithm. Then stacking is used to combine the multiple classifiers generated by the different algorithms identifying the best mix of classifiers.

Other approaches combine pre-processing techniques with bagging and boosting, simultaneously, composing an ensemble of ensembles. EasyEnsemble and BalanceCascade algorithms~\citep{liu2009exploratory} are examples of this type of approach. Both algorithms use bagging as the main ensemble method and use Adaboost for training each bag. As for the pre-processing technique, both construct balanced bags by randomly under-sampling examples from the majority class. In EasyEnsemble algorithm all Adaboost iterations can be performed simultaneously because each Adaboost ensemble uses a previously determined subset of the data. All the generated classifiers are combined for a final solution. 
 On the other hand, in the BalanceCascade algorithm, after the Adaboost learning, the majority examples correctly classified with higher confidence are discarded from further iterations.

\citet{wang2008combination} presents an approach that combines the SMOTE algorithm with Biased-SVM \citep{veropoulos1999controlling}. The proposed approach applies the Biased-SVM in the imbalanced data and stores the obtained support vectors from both classes. Then SMOTE is used to over-sample the support vectors with two alternatives: only use the obtained support vectors or use the entire minority class. A final classification is obtained with the new data using the biased-SVM.

Finally, a strategy using a clustering method based on class purity maximization is proposed by \citet{yoon2005unsupervised}. This method generates clusters of pure majority class examples and non-pure clusters based on the improvement of the clusters class purity. When the clusters are formed, all minority class examples are added to the non-pure clusters and a decision tree is built for each cluster. An unlabelled example is clustered according to the same algorithm. If it falls on a non-pure cluster, the decision tree committee votes the prediction, but if it falls on a pure majority class cluster the final prediction is  the majority class.
 If the committee votes for a majority class prediction, then that will be the final prediction. Onn the other hand, if it is a minority class prediction, then the example will be submitted to a final classifier which is constructed using a neural network.

\section{Related Problems}
\label{relatedprob}



In this section we describe some problems that frequently coexist with imbalanced data distributions and further contribute to degrade the performance of predictive models.
These related problems have been addressed mainly within a classification setting. 
Problems such as small disjuncts, class overlap and small sample size, usually coexist with imbalanced classification domains and are also identified as possible causes of classifiers performance degradation \citep{W04, he2009learning, sun2009classification}. We will briefly describe the major developments made for the following  related problems: class overlapping or class separability, 
small sample size and lack of density in the training set, high dimensionality of the data set, noisy data and small disjuncts.

The overlap problem occurs when a given region of the data space contains an identical number of training cases for each class. In this situation, a learner will have an increased difficulty in distinguishing between the classes present on the overlapping region. The problems of imbalanced data sets and overlapping regions were mostly treated separately. However, in the last decade, some attention was given to the relationship between these two problems  \citep{prati2004class, garcia2006combined}. The combination of imbalanced domains with overlapping regions causes an important deterioration of the learner performance and both problems acting together produce much more difficulties than expected when considering their effects individually \citep{denil2010overlap}.  Recent works \citep{alejo2011back, alejo2013hybrid} presented combinations of solutions for handling, simultaneously, both the class imbalance and the class overlap problem and apply a blend of techniques for addressing these issues. 

The small training set, or small sample problem, is also naturally related with imbalanced domains. In effect, having too few examples from the minority class 
will prevent the learner from  capturing their characteristics and will hinder the generalisation capability of the algorithm. 
The relation between imbalanced domains and small sample problems was addressed by \citet{japkowicz2002class} and  \citet{jo2004class}, where it was highlighted that class imbalance degrades classification performance in small data sets although this loss of performance tends to gradually reduce as the training set size increases. As expected, the subconcepts defined by the minority class examples can be better learned if their number can be increased.


The small sample problem may trigger problems such as rare cases \citep{weiss2005mining}, which bring an additional difficulty to the learning system. Rare examples are extremely scarce cases that are  difficult to detect and use for generalisation. The small training set problem may also be accompanied by a variable class distribution that may not match the target distribution.
\citet{forman2004learning} showed that, for imbalanced domains, obtaining a balanced training set is not the most favourable setting and classifiers performance can be greatly improved by non-random sampling that favours the minority class.

In some domains, such as text classification, the imbalance problem coexists with high dimensional data sets, i.e., domains with a high number of predictors.
The main challenge here is to adequately select features that contain the key information of the problem.  Feature selection is recommended \citep{wasikowski2010combating} and is also pointed as the solution for addressing the class imbalance problem \citep{mladenic1999feature, zheng2004feature, chen2008fast, van2004bias, forman2003extensive}.
Several proposals exist for handling the imbalance problem in conjunction with the high dimensionality problem, all using a feature selection strategy \citep{zheng2004feature, del2004multistrategy, forman2004learning, chu2010new}. 

Noise is a known factor that usually affects models performance. In imbalanced domains, noisy data has a greater impact on the least represented examples \citep{W04}. A recent study  \citep{seiffert2011empirical} on the effect of noise in a data set intrinsically characterised by the presence of both class imbalance and class noise concluded that, generally, class noise has a more significant impact on learners than imbalance. It was also noticed that the interaction between the level of imbalance and the level of noise within a data set is a significant factor, and that studying these two cal effects  individually may not be sufficient.

One of the most studied related problems is the problem of small disjuncts which is associated to the imbalance in the subclusters of each class in the data set \citep{japkowicz2001concept, jo2004class}. When a subcluster has a low \textit{coverage}, i.e., it classifies few examples, it is called small~\citep{holte1989concept}.
Small disjuncts are a problem because the learners are typically biased towards classifying large disjuncts and therefore they will tend to overfit and misclassify the cases in the small disjuncts.
This problem is often present along with the problem of class imbalance in real world data sets and the connection  between the two problems is not yet well understood \cite{jo2004class}.
Due to the importance of these two problems, several works address the relation between the problem of small disjuncts and the class imbalance problem~\citep{japkowicz2003class, weiss2003learning,jo2004class,pearson2003imbalanced, japkowicz2001concept, prati2004learning}.
A new metric called \textit{error concentration}~\citep{weiss2000quantitative} was proposed for evaluating the error concentration towards the smaller disjuncts. The work in \citet{weiss2010impact} analyses the impact of several factors on small disjuncts and in the error distribution across disjuncts. Among the studied factors are pruning, training-set size, noise and class imbalance.
Regarding pruning, it was not considered an effective strategy for dealing with small disjuncts in the presence of class imbalance~\citep{prati2004learning, weiss2010impact}. 
\citet{weiss2010impact} also concluded that even with a balanced data set, errors tend to be concentrated towards the smaller disjuncts. However, when there is class imbalance, the error concentration increases. Moreover, the increase in the class imbalance also increases the error concentration.
Thus, class imbalance is partly responsible for the problem with small disjuncts, and artificially modifying the class distribution of the training data to be more balanced, causes a decrease in the error concentration.

All the considered problems coexist and are related with the imbalance problem. The conjunction of these problems with imbalanced domains tends to further degrade the classifiers performance and therefore this relationship should not be ignored.

\section{Conclusions}
\label{conc}

Imbalanced domains pose important challenges to existing approaches to predictive modelling. In this paper we propose a  formulation of the problem of modelling using imbalanced data sets that includes both classification and regression tasks. We present a survey of the state of the art solutions  for obtaining and evaluating predictive models for both classification and regression tasks. We propose a new taxonomy for the existing approaches grouping them into: (i)  data pre-processing, (ii) special-purpose learning methods and (iii) prediction post-processing.

Most existing solutions to modelling under imbalanced distributions are focused on classification tasks. This fact is also present on previous surveys of this important research area. In this paper, we propose the first survey that also addresses existing approaches to imbalanced data sets within regression tasks.

Finally, we describe some problems that are strongly related with imbalanced data distributions, highlighting works that explore the relationship of these other problems with  imbalance data sets.




\bibliographystyle{apa}
\bibliography{revisedBib.bib}

\begin{thebibliography}{}

\bibitem[\protect\astroncite{Akbani et~al.}{2004}]{akbani2004applying}
Akbani, R., Kwek, S., and Japkowicz, N. (2004).
\newblock Applying support vector machines to imbalanced datasets.
\newblock In {\em Machine Learning: ECML 2004}, pages 39--50. Springer.

\bibitem[\protect\astroncite{Alejo et~al.}{2007}]{alejo2007improving}
Alejo, R., Garc{\'\i}a, V., Sotoca, J.~M., Mollineda, R.~A., and S{\'a}nchez,
  J.~S. (2007).
\newblock Improving the performance of the rbf neural networks trained with
  imbalanced samples.
\newblock In {\em Computational and Ambient Intelligence}, pages 162--169.
  Springer.

\bibitem[\protect\astroncite{Alejo et~al.}{2013}]{alejo2013hybrid}
Alejo, R., Valdovinos, R.~M., Garc{\'\i}a, V., and Pacheco-Sanchez, J. (2013).
\newblock A hybrid method to face class overlap and class imbalance on neural
  networks and multi-class scenarios.
\newblock {\em Pattern Recognition Letters}, 34(4):380--388.

\bibitem[\protect\astroncite{Alejo~Eleuterio et~al.}{2011}]{alejo2011back}
Alejo~Eleuterio, R., Mart{\'\i}nez~Sotoca, J., Garc{\'\i}a~Jim{\'e}nez, V., and
  Valdovinos~Rosas, R.~M. (2011).
\newblock Back propagation with balanced mse cost function and nearest neighbor
  editing for handling class overlap and class imbalance.

\bibitem[\protect\astroncite{Bansal et~al.}{2008}]{bansal2008tuning}
Bansal, G., Sinha, A.~P., and Zhao, H. (2008).
\newblock Tuning data mining methods for cost-sensitive regression: a study in
  loan charge-off forecasting.
\newblock {\em Journal of Management Information Systems}, 25(3):315--336.

\bibitem[\protect\astroncite{Barandela et~al.}{2003}]{barandela2003strategies}
Barandela, R., S{\'a}nchez, J.~S., Garc{\i}a, V., and Rangel, E. (2003).
\newblock Strategies for learning in class imbalance problems.
\newblock {\em Pattern Recognition}, 36(3):849--851.

\bibitem[\protect\astroncite{Barua et~al.}{2012}]{barua2012mwmote}
Barua, S., Islam, M., Yao, X., and Murase, K. (2012).
\newblock Mwmote-majority weighted minority oversampling technique for
  imbalanced data set learning.

\bibitem[\protect\astroncite{Batista et~al.}{2004}]{batista2004study}
Batista, G.~E., Prati, R.~C., and Monard, M.~C. (2004).
\newblock A study of the behavior of several methods for balancing machine
  learning training data.
\newblock {\em ACM SIGKDD Explorations Newsletter}, 6(1):20--29.

\bibitem[\protect\astroncite{Batuwita and Palade}{2009}]{batuwita2009new}
Batuwita, R. and Palade, V. (2009).
\newblock A new performance measure for class imbalance learning. application
  to bioinformatics problems.
\newblock In {\em Machine Learning and Applications, 2009. ICMLA'09.
  International Conference on}, pages 545--550. IEEE.

\bibitem[\protect\astroncite{Batuwita and
  Palade}{2010a}]{batuwita2010efficient}
Batuwita, R. and Palade, V. (2010a).
\newblock Efficient resampling methods for training support vector machines
  with imbalanced datasets.
\newblock In {\em Neural Networks (IJCNN), The 2010 International Joint
  Conference on}, pages 1--8. IEEE.

\bibitem[\protect\astroncite{Batuwita and Palade}{2010b}]{batuwita2010fsvm}
Batuwita, R. and Palade, V. (2010b).
\newblock Fsvm-cil: fuzzy support vector machines for class imbalance learning.
\newblock {\em Fuzzy Systems, IEEE Transactions on}, 18(3):558--571.

\bibitem[\protect\astroncite{Batuwita and Palade}{2012}]{batuwita2012adjusted}
Batuwita, R. and Palade, V. (2012).
\newblock Adjusted geometric-mean: a novel performance measure for imbalanced
  bioinformatics datasets learning.
\newblock {\em Journal of Bioinformatics and Computational Biology}, 10(04).

\bibitem[\protect\astroncite{Bellinger et~al.}{2012}]{bellinger2012one}
Bellinger, C., Sharma, S., and Japkowicz, N. (2012).
\newblock One-class versus binary classification: Which and when?
\newblock In {\em Machine Learning and Applications (ICMLA), 2012 11th
  International Conference on}, volume~2, pages 102--106. IEEE.

\bibitem[\protect\astroncite{Bi and Bennett}{2003}]{BB03}
Bi, J. and Bennett, K.~P. (2003).
\newblock Regression error characteristic curves.
\newblock In {\em Proc. of the 20th Int. Conf. on Machine Learning}, pages
  43--50.

\bibitem[\protect\astroncite{Bradley}{1997}]{bradley1997use}
Bradley, A.~P. (1997).
\newblock The use of the area under the roc curve in the evaluation of machine
  learning algorithms.
\newblock {\em Pattern recognition}, 30(7):1145--1159.

\bibitem[\protect\astroncite{Breiman et~al.}{1984}]{breiman1984classification}
Breiman, L., Friedman, J.~H., Olshen, R.~A., and Stone, C.~J. (1984).
\newblock Classification and regression trees. wadsworth \& brooks.
\newblock {\em Monterey, CA}.

\bibitem[\protect\astroncite{Bunkhumpornpat
  et~al.}{2009}]{bunkhumpornpat2009safe}
Bunkhumpornpat, C., Sinapiromsaran, K., and Lursinsap, C. (2009).
\newblock Safe-level-smote: Safe-level-synthetic minority over-sampling
  technique for handling the class imbalanced problem.
\newblock In {\em Advances in Knowledge Discovery and Data Mining}, pages
  475--482. Springer.

\bibitem[\protect\astroncite{Bunkhumpornpat
  et~al.}{2011}]{bunkhumpornpat2011mute}
Bunkhumpornpat, C., Sinapiromsaran, K., and Lursinsap, C. (2011).
\newblock Mute: Majority under-sampling technique.
\newblock In {\em Information, Communications and Signal Processing (ICICS)
  2011 8th International Conference on}, pages 1--4. IEEE.

\bibitem[\protect\astroncite{Bunkhumpornpat
  et~al.}{2012}]{bunkhumpornpat2012dbsmote}
Bunkhumpornpat, C., Sinapiromsaran, K., and Lursinsap, C. (2012).
\newblock Dbsmote: Density-based synthetic minority over-sampling technique.
\newblock {\em Applied Intelligence}, 36(3):664--684.

\bibitem[\protect\astroncite{Bunkhumpornpat and
  Subpaiboonkit}{2013}]{bunkhumpornpat2013safe}
Bunkhumpornpat, C. and Subpaiboonkit, S. (2013).
\newblock Safe level graph for synthetic minority over-sampling techniques.
\newblock In {\em Communications and Information Technologies (ISCIT), 2013
  13th International Symposium on}, pages 570--575. IEEE.

\bibitem[\protect\astroncite{Cain and Janssen}{1995}]{cain1995real}
Cain, M. and Janssen, C. (1995).
\newblock Real estate price prediction under asymmetric loss.
\newblock {\em Annals of the Institute of Statistical Mathematics},
  47(3):401--414.

\bibitem[\protect\astroncite{Cao et~al.}{2013}]{cao2013pso}
Cao, P., Zhao, D., and Za{\"\i}ane, O.~R. (2013).
\newblock A pso-based cost-sensitive neural network for imbalanced data
  classification.
\newblock In {\em Trends and Applications in Knowledge Discovery and Data
  Mining}, pages 452--463. Springer.

\bibitem[\protect\astroncite{Castro and
  de~P{\'a}dua~Braga}{2013}]{castro2013novel}
Castro, C.~L. and de~P{\'a}dua~Braga, A. (2013).
\newblock Novel cost-sensitive approach to improve the multilayer perceptron
  performance on imbalanced data.
\newblock {\em IEEE Trans. Neural Netw. Learning Syst.}, 24(6):888--899.

\bibitem[\protect\astroncite{Chang et~al.}{2003}]{chang2003statistical}
Chang, E.~Y., Li, B., Wu, G., and Goh, K. (2003).
\newblock Statistical learning for effective visual information retrieval.
\newblock In {\em ICIP (3)}, pages 609--612.

\bibitem[\protect\astroncite{Chawla et~al.}{2002}]{CBOK02}
Chawla, N.~V., Bowyer, K.~W., Hall, L.~O., and Kegelmeyer, W.~P. (2002).
\newblock Smote: Synthetic minority over-sampling technique.
\newblock {\em JAIR}, 16:321--357.

\bibitem[\protect\astroncite{Chawla et~al.}{2008}]{chawla2008automatically}
Chawla, N.~V., Cieslak, D.~A., Hall, L.~O., and Joshi, A. (2008).
\newblock Automatically countering imbalance and its empirical relationship to
  cost.
\newblock {\em Data Mining and Knowledge Discovery}, 17(2):225--252.

\bibitem[\protect\astroncite{Chawla et~al.}{2005}]{chawla2005wrapper}
Chawla, N.~V., Hall, L.~O., and Joshi, A. (2005).
\newblock Wrapper-based computation and evaluation of sampling methods for
  imbalanced datasets.
\newblock In {\em Proceedings of the 1st international workshop on
  Utility-based data mining}, pages 24--33. ACM.

\bibitem[\protect\astroncite{Chawla et~al.}{2004}]{chawla2004editorial}
Chawla, N.~V., Japkowicz, N., and Kotcz, A. (2004).
\newblock Editorial: special issue on learning from imbalanced data sets.
\newblock {\em ACM SIGKDD Explorations Newsletter}, 6(1):1--6.

\bibitem[\protect\astroncite{Chawla et~al.}{2003}]{chawla2003smoteboost}
Chawla, N.~V., Lazarevic, A., Hall, L.~O., and Bowyer, K.~W. (2003).
\newblock Smoteboost: Improving prediction of the minority class in boosting.
\newblock In {\em Knowledge Discovery in Databases: PKDD 2003}, pages 107--119.
  Springer.

\bibitem[\protect\astroncite{Chen et~al.}{2004}]{chen2004using}
Chen, C., Liaw, A., and Breiman, L. (2004).
\newblock Using random forest to learn imbalanced data.
\newblock {\em University of California, Berkeley}.

\bibitem[\protect\astroncite{Chen et~al.}{2010}]{chen2010ramoboost}
Chen, S., He, H., and Garcia, E.~A. (2010).
\newblock Ramoboost: Ranked minority oversampling in boosting.
\newblock {\em Neural Networks, IEEE Transactions on}, 21(10):1624--1642.

\bibitem[\protect\astroncite{Chen and Wasikowski}{2008}]{chen2008fast}
Chen, X.-w. and Wasikowski, M. (2008).
\newblock Fast: a roc-based feature selection metric for small samples and
  imbalanced data classification problems.
\newblock In {\em Proceedings of the 14th ACM SIGKDD international conference
  on Knowledge discovery and data mining}, pages 124--132. ACM.

\bibitem[\protect\astroncite{Christoffersen and
  Diebold}{1996}]{christoffersen1996further}
Christoffersen, P.~F. and Diebold, F.~X. (1996).
\newblock Further results on forecasting and model selection under asymmetric
  loss.
\newblock {\em Journal of applied econometrics}, 11(5):561--571.

\bibitem[\protect\astroncite{Christoffersen and
  Diebold}{1997}]{christoffersen1997optimal}
Christoffersen, P.~F. and Diebold, F.~X. (1997).
\newblock Optimal prediction under asymmetric loss.
\newblock {\em Econometric theory}, 13(06):808--817.

\bibitem[\protect\astroncite{Chu et~al.}{2010}]{chu2010new}
Chu, L., Gao, H., and Chang, W. (2010).
\newblock A new feature weighting method based on probability distribution in
  imbalanced text classification.
\newblock In {\em Fuzzy Systems and Knowledge Discovery (FSKD), 2010 Seventh
  International Conference on}, volume~5, pages 2335--2339. IEEE.

\bibitem[\protect\astroncite{Chyi}{2003}]{Chyi2003class}
Chyi, Y.-M. (2003).
\newblock Classification analysis techniques for skewed class distribution
  problems.
\newblock {\em Master Thesis, Department of Information Management, National
  Sun Yat-Sen University}.

\bibitem[\protect\astroncite{Cieslak and Chawla}{2008}]{cieslak2008learning}
Cieslak, D.~A. and Chawla, N.~V. (2008).
\newblock Learning decision trees for unbalanced data.
\newblock In {\em Machine Learning and Knowledge Discovery in Databases}, pages
  241--256. Springer.

\bibitem[\protect\astroncite{Cieslak et~al.}{2012}]{cieslak2012hellinger}
Cieslak, D.~A., Hoens, T.~R., Chawla, N.~V., and Kegelmeyer, W.~P. (2012).
\newblock Hellinger distance decision trees are robust and skew-insensitive.
\newblock {\em Data Mining and Knowledge Discovery}, 24(1):136--158.

\bibitem[\protect\astroncite{Cohen et~al.}{2006}]{cohen2006learning}
Cohen, G., Hilario, M., Sax, H., Hugonnet, S., and Geissbuhler, A. (2006).
\newblock Learning from imbalanced data in surveillance of nosocomial
  infection.
\newblock {\em Artificial Intelligence in Medicine}, 37(1):7--18.

\bibitem[\protect\astroncite{Crone et~al.}{2005}]{crone2005utility}
Crone, S.~F., Lessmann, S., and Stahlbock, R. (2005).
\newblock Utility based data mining for time series analysis: cost-sensitive
  learning for neural network predictors.
\newblock In {\em Proceedings of the 1st international workshop on
  Utility-based data mining}, pages 59--68. ACM.

\bibitem[\protect\astroncite{Daskalaki et~al.}{2006}]{daskalaki2006evaluation}
Daskalaki, S., Kopanas, I., and Avouris, N.~M. (2006).
\newblock Evaluation of classifiers for an uneven class distribution problem.
\newblock {\em Applied Artificial Intelligence}, 20(5):381--417.

\bibitem[\protect\astroncite{Davis and Goadrich}{2006}]{DG06}
Davis, J. and Goadrich, M. (2006).
\newblock The relationship between precision-recall and roc curves.
\newblock In {\em ICML'06: Proc. of the 23rd Int. Conf. on Machine Learning},
  ACM ICPS, pages 233--240. ACM.

\bibitem[\protect\astroncite{Del~Castillo and
  Serrano}{2004}]{del2004multistrategy}
Del~Castillo, M.~D. and Serrano, J.~I. (2004).
\newblock A multistrategy approach for digital text categorization from
  imbalanced documents.
\newblock {\em ACM SIGKDD Explorations Newsletter}, 6(1):70--79.

\bibitem[\protect\astroncite{Denil and Trappenberg}{2010}]{denil2010overlap}
Denil, M. and Trappenberg, T. (2010).
\newblock Overlap versus imbalance.
\newblock In {\em Advances in Artificial Intelligence}, pages 220--231.
  Springer.

\bibitem[\protect\astroncite{Domingos}{1999}]{Dom99}
Domingos, P. (1999).
\newblock Metacost: A general method for making classifiers cost-sensitive.
\newblock In {\em KDD'99: Proceedings of the 5th International Conference on
  Knowledge Discovery and Data Mining}, pages 155--164. ACM Press.

\bibitem[\protect\astroncite{Doucette and Heywood}{2008}]{doucette2008gp}
Doucette, J. and Heywood, M.~I. (2008).
\newblock Gp classification under imbalanced data sets: Active sub-sampling and
  auc approximation.
\newblock In {\em Genetic Programming}, pages 266--277. Springer.

\bibitem[\protect\astroncite{Drown et~al.}{2009}]{drown2009evolutionary}
Drown, D.~J., Khoshgoftaar, T.~M., and Seliya, N. (2009).
\newblock Evolutionary sampling and software quality modeling of high-assurance
  systems.
\newblock {\em Systems, Man and Cybernetics, Part A: Systems and Humans, IEEE
  Transactions on}, 39(5):1097--1107.

\bibitem[\protect\astroncite{Drummond and Holte}{2003}]{drummond2003c4}
Drummond, C. and Holte, R.~C. (2003).
\newblock C4. 5, class imbalance, and cost sensitivity: why under-sampling
  beats over-sampling.
\newblock In {\em Workshop on Learning from Imbalanced Datasets II}, volume~11.
  Citeseer.

\bibitem[\protect\astroncite{Egan}{1975}]{egan1975signal}
Egan, J.~P. (1975).
\newblock Signal detection theory and $\{$ROC$\}$ analysis.

\bibitem[\protect\astroncite{Ertekin}{2013}]{ertekin2013virtual}
Ertekin, {\c{S}}. (2013).
\newblock Adaptive oversampling for imbalanced data classification.
\newblock In {\em Information Sciences and Systems 2013}, pages 261--269.
  Springer.

\bibitem[\protect\astroncite{Ertekin et~al.}{2007a}]{ertekin2007learning}
Ertekin, {\c{S}}., Huang, J., Bottou, L., and Giles, L. (2007a).
\newblock Learning on the border: active learning in imbalanced data
  classification.
\newblock In {\em Proceedings of the sixteenth ACM conference on Conference on
  information and knowledge management}, pages 127--136. ACM.

\bibitem[\protect\astroncite{Ertekin et~al.}{2007b}]{ertekin2007active}
Ertekin, {\c{S}}., Huang, J., and Giles, C.~L. (2007b).
\newblock Active learning for class imbalance problem.
\newblock In {\em Proceedings of the 30th annual international ACM SIGIR
  conference on Research and development in information retrieval}, pages
  823--824. ACM.

\bibitem[\protect\astroncite{Estabrooks and
  Japkowicz}{2001}]{estabrooks2001mixture}
Estabrooks, A. and Japkowicz, N. (2001).
\newblock A mixture-of-experts framework for learning from imbalanced data
  sets.
\newblock In {\em Advances in Intelligent Data Analysis}, pages 34--43.
  Springer.

\bibitem[\protect\astroncite{Estabrooks et~al.}{2004}]{estabrooks2004multiple}
Estabrooks, A., Jo, T., and Japkowicz, N. (2004).
\newblock A multiple resampling method for learning from imbalanced data sets.
\newblock {\em Computational Intelligence}, 20(1):18--36.

\bibitem[\protect\astroncite{Fern{\'a}ndez et~al.}{2010}]{fernandez20102}
Fern{\'a}ndez, A., del Jesus, M.~J., and Herrera, F. (2010).
\newblock On the 2-tuples based genetic tuning performance for fuzzy rule based
  classification systems in imbalanced data-sets.
\newblock {\em Information Sciences}, 180(8):1268--1291.

\bibitem[\protect\astroncite{Fern{\'a}ndez et~al.}{2008}]{fernandez2008study}
Fern{\'a}ndez, A., Garc{\'\i}a, S., del Jesus, M.~J., and Herrera, F. (2008).
\newblock A study of the behaviour of linguistic fuzzy rule based
  classification systems in the framework of imbalanced data-sets.
\newblock {\em Fuzzy Sets and Systems}, 159(18):2378--2398.

\bibitem[\protect\astroncite{Forman}{2003}]{forman2003extensive}
Forman, G. (2003).
\newblock An extensive empirical study of feature selection metrics for text
  classification.
\newblock {\em The Journal of machine learning research}, 3:1289--1305.

\bibitem[\protect\astroncite{Forman and Cohen}{2004}]{forman2004learning}
Forman, G. and Cohen, I. (2004).
\newblock Learning from little: Comparison of classifiers given little
  training.
\newblock In {\em Knowledge Discovery in Databases: PKDD 2004}, pages 161--172.
  Springer.

\bibitem[\protect\astroncite{Galar et~al.}{2012}]{galar2012review}
Galar, M., Fern{\'a}ndez, A., Barrenechea, E., Bustince, H., and Herrera, F.
  (2012).
\newblock A review on ensembles for the class imbalance problem: bagging-,
  boosting-, and hybrid-based approaches.
\newblock {\em Systems, Man, and Cybernetics, Part C: Applications and Reviews,
  IEEE Transactions on}, 42(4):463--484.

\bibitem[\protect\astroncite{Galar et~al.}{2013}]{galar2013eusboost}
Galar, M., Fern{\'a}ndez, A., Barrenechea, E., and Herrera, F. (2013).
\newblock Eusboost: Enhancing ensembles for highly imbalanced data-sets by
  evolutionary undersampling.
\newblock {\em Pattern Recognition}.

\bibitem[\protect\astroncite{Garc{\'i}a et~al.}{2012}]{derrac2012evolutionary}
Garc{\'i}a, Salvador~Derrac, J., Triguero, I., Carmona, C.~J., and Herrera, F.
  (2012).
\newblock Evolutionary-based selection of generalized instances for imbalanced
  classification.
\newblock {\em Knowledge-Based Systems}, 25(1):3--12.

\bibitem[\protect\astroncite{Garc{\'\i}a et~al.}{2006a}]{garcia2006proposal}
Garc{\'\i}a, S., Cano, J.~R., Fern{\'a}ndez, A., and Herrera, F. (2006a).
\newblock A proposal of evolutionary prototype selection for class imbalance
  problems.
\newblock In {\em Intelligent Data Engineering and Automated Learning--IDEAL
  2006}, pages 1415--1423. Springer.

\bibitem[\protect\astroncite{Garc{\'\i}a and
  Herrera}{2009}]{garcia2009evolutionary}
Garc{\'\i}a, S. and Herrera, F. (2009).
\newblock Evolutionary undersampling for classification with imbalanced
  datasets: Proposals and taxonomy.
\newblock {\em Evolutionary Computation}, 17(3):275--306.

\bibitem[\protect\astroncite{Garc{\'\i}a et~al.}{2006b}]{garcia2006combined}
Garc{\'\i}a, V., Alejo, R., S{\'a}nchez, J.~S., Sotoca, J.~M., and Mollineda,
  R.~A. (2006b).
\newblock Combined effects of class imbalance and class overlap on
  instance-based classification.
\newblock In {\em Intelligent Data Engineering and Automated Learning--IDEAL
  2006}, pages 371--378. Springer.

\bibitem[\protect\astroncite{Garc{\'\i}a et~al.}{2008}]{garcia2008new}
Garc{\'\i}a, V., Mollineda, R.~A., and S{\'a}nchez, J.~S. (2008).
\newblock A new performance evaluation method for two-class imbalanced
  problems.
\newblock In {\em Structural, Syntactic, and Statistical Pattern Recognition},
  pages 917--925. Springer.

\bibitem[\protect\astroncite{Garc{\'\i}a et~al.}{2009}]{garcia2009index}
Garc{\'\i}a, V., Mollineda, R.~A., and S{\'a}nchez, J.~S. (2009).
\newblock Index of balanced accuracy: A performance measure for skewed class
  distributions.
\newblock In {\em Pattern Recognition and Image Analysis}, pages 441--448.
  Springer.

\bibitem[\protect\astroncite{Garc{\'\i}a et~al.}{2010}]{garcia2010theoretical}
Garc{\'\i}a, V., Mollineda, R.~A., and S{\'a}nchez, J.~S. (2010).
\newblock Theoretical analysis of a performance measure for imbalanced data.
\newblock In {\em Pattern Recognition (ICPR), 2010 20th International
  Conference on}, pages 617--620. IEEE.

\bibitem[\protect\astroncite{Granger}{1999}]{granger1999outline}
Granger, C.~W. (1999).
\newblock Outline of forecast theory using generalized cost functions.
\newblock {\em Spanish Economic Review}, 1(2):161--173.

\bibitem[\protect\astroncite{Han et~al.}{2005}]{han2005borderline}
Han, H., Wang, W.-Y., and Mao, B.-H. (2005).
\newblock Borderline-smote: A new over-sampling method in imbalanced data sets
  learning.
\newblock In {\em Advances in intelligent computing}, pages 878--887. Springer.

\bibitem[\protect\astroncite{Hand}{2009}]{hand2009measuring}
Hand, D.~J. (2009).
\newblock Measuring classifier performance: a coherent alternative to the area
  under the roc curve.
\newblock {\em Machine learning}, 77(1):103--123.

\bibitem[\protect\astroncite{Hart}{1968}]{cnn}
Hart, P.~E. (1968).
\newblock The condensed nearest neighbor rule.
\newblock {\em IEEE Transactions on Information Theory}, 14:515--516.

\bibitem[\protect\astroncite{He et~al.}{2008}]{he2008adasyn}
He, H., Bai, Y., Garcia, E.~A., and Li, S. (2008).
\newblock Adasyn: Adaptive synthetic sampling approach for imbalanced learning.
\newblock In {\em Neural Networks, 2008. IJCNN 2008.(IEEE World Congress on
  Computational Intelligence). IEEE International Joint Conference on}, pages
  1322--1328. IEEE.

\bibitem[\protect\astroncite{He and Garcia}{2009}]{he2009learning}
He, H. and Garcia, E.~A. (2009).
\newblock Learning from imbalanced data.
\newblock {\em Knowledge and Data Engineering, IEEE Transactions on},
  21(9):1263--1284.

\bibitem[\protect\astroncite{Hern{\'a}ndez-Orallo}{2012}]{hernandez2012soft}
Hern{\'a}ndez-Orallo, J. (2012).
\newblock Soft (gaussian cde) regression models and loss functions.
\newblock {\em arXiv preprint arXiv:1211.1043}.

\bibitem[\protect\astroncite{Hern{\'a}ndez-Orallo}{2013}]{HernandezOrallo20133395}
Hern{\'a}ndez-Orallo, J. (2013).
\newblock \{ROC\} curves for regression.
\newblock {\em Pattern Recognition}, 46(12):3395 -- 3411.

\bibitem[\protect\astroncite{Hern{\'a}ndez-Orallo}{2014}]{orallo2014prob}
Hern{\'a}ndez-Orallo, J. (2014).
\newblock Probabilistic reframing for cost-sensitive regression.
\newblock {\em ACM Trans. Knowl. Discov. Data}, 8(4):17:1--17:55.

\bibitem[\protect\astroncite{Holte et~al.}{1989}]{holte1989concept}
Holte, R.~C., Acker, L.~E., and Porter, B.~W. (1989).
\newblock Concept learning and the problem of small disjuncts.
\newblock In {\em IJCAI}, volume~89, pages 813--818. Citeseer.

\bibitem[\protect\astroncite{Hu}{2012}]{hu2012active}
Hu, J. (2012).
\newblock Active learning for imbalance problem using l-gem of rbfnn.
\newblock In {\em ICMLC}, pages 490--495.

\bibitem[\protect\astroncite{Hu et~al.}{2009}]{hu2009msmote}
Hu, S., Liang, Y., Ma, L., and He, Y. (2009).
\newblock Msmote: improving classification performance when training data is
  imbalanced.
\newblock In {\em Computer Science and Engineering, 2009. WCSE'09. Second
  International Workshop on}, volume~2, pages 13--17. IEEE.

\bibitem[\protect\astroncite{Huang et~al.}{2004}]{huang2004learning}
Huang, K., Yang, H., King, I., and Lyu, M.~R. (2004).
\newblock Learning classifiers from imbalanced data based on biased minimax
  probability machine.
\newblock In {\em Computer Vision and Pattern Recognition, 2004. CVPR 2004.
  Proceedings of the 2004 IEEE Computer Society Conference on}, volume~2, pages
  II--558. IEEE.

\bibitem[\protect\astroncite{Hwang et~al.}{2011}]{hwang2011new}
Hwang, J.~P., Park, S., and Kim, E. (2011).
\newblock A new weighted approach to imbalanced data classification problem via
  support vector machine with quadratic cost function.
\newblock {\em Expert Systems with Applications}, 38(7):8580--8585.

\bibitem[\protect\astroncite{Imam et~al.}{2006}]{imam2006z}
Imam, T., Ting, K.~M., and Kamruzzaman, J. (2006).
\newblock z-svm: An svm for improved classification of imbalanced data.
\newblock In {\em AI 2006: Advances in Artificial Intelligence}, pages
  264--273. Springer.

\bibitem[\protect\astroncite{Japkowicz}{2000}]{japkowicz2000learning}
Japkowicz, N. (2000).
\newblock Learning from imbalanced data sets: a comparison of various
  strategies.
\newblock In {\em AAAI workshop on learning from imbalanced data sets},
  volume~68. Menlo Park, CA.

\bibitem[\protect\astroncite{Japkowicz}{2001}]{japkowicz2001concept}
Japkowicz, N. (2001).
\newblock Concept-learning in the presence of between-class and within-class
  imbalances.
\newblock In {\em Advances in Artificial Intelligence}, pages 67--77. Springer.

\bibitem[\protect\astroncite{Japkowicz}{2003}]{japkowicz2003class}
Japkowicz, N. (2003).
\newblock Class imbalances: are we focusing on the right issue.
\newblock In {\em Workshop on Learning from Imbalanced Data Sets II}, volume
  1723, page~63.

\bibitem[\protect\astroncite{Japkowicz et~al.}{1995}]{japkowicz1995novelty}
Japkowicz, N., Myers, C., and Gluck, M. (1995).
\newblock A novelty detection approach to classification.
\newblock In {\em IJCAI}, pages 518--523.

\bibitem[\protect\astroncite{Japkowicz and Stephen}{2002}]{japkowicz2002class}
Japkowicz, N. and Stephen, S. (2002).
\newblock The class imbalance problem: A systematic study.
\newblock {\em Intelligent data analysis}, 6(5):429--449.

\bibitem[\protect\astroncite{Jeatrakul
  et~al.}{2010}]{jeatrakul2010classification}
Jeatrakul, P., Wong, K.~W., and Fung, C.~C. (2010).
\newblock Classification of imbalanced data by combining the complementary
  neural network and smote algorithm.
\newblock In {\em Neural Information Processing. Models and Applications},
  pages 152--159. Springer.

\bibitem[\protect\astroncite{Jo and Japkowicz}{2004}]{jo2004class}
Jo, T. and Japkowicz, N. (2004).
\newblock Class imbalances versus small disjuncts.
\newblock {\em ACM SIGKDD Explorations Newsletter}, 6(1):40--49.

\bibitem[\protect\astroncite{Joshi et~al.}{2001}]{joshi2001evaluating}
Joshi, M.~V., Kumar, V., and Agarwal, R.~C. (2001).
\newblock Evaluating boosting algorithms to classify rare classes: Comparison
  and improvements.
\newblock In {\em Data Mining, 2001. ICDM 2001, Proceedings IEEE International
  Conference on}, pages 257--264. IEEE.

\bibitem[\protect\astroncite{Kang and Cho}{2006}]{kang2006eus}
Kang, P. and Cho, S. (2006).
\newblock Eus svms: Ensemble of under-sampled svms for data imbalance problems.
\newblock In {\em Neural Information Processing}, pages 837--846. Springer.

\bibitem[\protect\astroncite{Kotsiantis et~al.}{2006}]{kotsiantis2006handling}
Kotsiantis, S., Kanellopoulos, D., and Pintelas, P. (2006).
\newblock Handling imbalanced datasets: A review.
\newblock {\em GESTS International Transactions on Computer Science and
  Engineering}, 30(1):25--36.

\bibitem[\protect\astroncite{Kotsiantis and
  Pintelas}{2003}]{kotsiantis2003mixture}
Kotsiantis, S. and Pintelas, P. (2003).
\newblock Mixture of expert agents for handling imbalanced data sets.
\newblock {\em Annals of Mathematics, Computing \& Teleinformatics},
  1(1):46--55.

\bibitem[\protect\astroncite{Kubat et~al.}{1998}]{kubat1998machine}
Kubat, M., Holte, R.~C., and Matwin, S. (1998).
\newblock Machine learning for the detection of oil spills in satellite radar
  images.
\newblock {\em Machine learning}, 30(2-3):195--215.

\bibitem[\protect\astroncite{Kubat and Matwin}{1997}]{KM97}
Kubat, M. and Matwin, S. (1997).
\newblock Addressing the curse of imbalanced training sets: One-sided
  selection.
\newblock In {\em Proc. of the 14th Int. Conf. on Machine Learning}, pages
  179--186. Morgan Kaufmann.

\bibitem[\protect\astroncite{Laurikkala}{2001}]{laurikkala2001improving}
Laurikkala, J. (2001).
\newblock {\em Improving identification of difficult small classes by balancing
  class distribution}.
\newblock Springer.

\bibitem[\protect\astroncite{Lee and Cho}{2006}]{lee2006novelty}
Lee, H.-j. and Cho, S. (2006).
\newblock The novelty detection approach for different degrees of class
  imbalance.
\newblock In {\em Neural Information Processing}, pages 21--30. Springer.

\bibitem[\protect\astroncite{Lee}{1999}]{lee1999regularization}
Lee, S.~S. (1999).
\newblock Regularization in skewed binary classification.
\newblock {\em Computational Statistics}, 14(2):277.

\bibitem[\protect\astroncite{Lee}{2000}]{lee2000noisy}
Lee, S.~S. (2000).
\newblock Noisy replication in skewed binary classification.
\newblock {\em Computational statistics \& data analysis}, 34(2):165--191.

\bibitem[\protect\astroncite{Lee}{2008}]{lee2008loss}
Lee, T.-H. (2008).
\newblock Loss functions in time series forecasting.
\newblock {\em International encyclopedia of the social sciences}.

\bibitem[\protect\astroncite{Li et~al.}{2009}]{li2009improved}
Li, C., Jing, C., and Xin-tao, G. (2009).
\newblock An improved p-svm method used to deal with imbalanced data sets.
\newblock In {\em Intelligent Computing and Intelligent Systems, 2009. ICIS
  2009. IEEE International Conference on}, volume~1, pages 118--122. IEEE.

\bibitem[\protect\astroncite{Li et~al.}{2008}]{li2008hybrid}
Li, P., Qiao, P.-L., and Liu, Y.-C. (2008).
\newblock A hybrid re-sampling method for svm learning from imbalanced data
  sets.
\newblock In {\em Fuzzy Systems and Knowledge Discovery, 2008. FSKD'08. Fifth
  International Conference on}, volume~2, pages 65--69. IEEE.

\bibitem[\protect\astroncite{Lin and Wang}{2002}]{lin2002fuzzy}
Lin, C.-F. and Wang, S.-D. (2002).
\newblock Fuzzy support vector machines.
\newblock {\em Neural Networks, IEEE Transactions on}, 13(2):464--471.

\bibitem[\protect\astroncite{Liu et~al.}{2007}]{liu2007generative}
Liu, A., Ghosh, J., and Martin, C.~E. (2007).
\newblock Generative oversampling for mining imbalanced datasets.
\newblock In {\em DMIN}, pages 66--72.

\bibitem[\protect\astroncite{Liu et~al.}{2010}]{liu2010robust}
Liu, W., Chawla, S., Cieslak, D.~A., and Chawla, N.~V. (2010).
\newblock A robust decision tree algorithm for imbalanced data sets.
\newblock In {\em SDM}, volume~10, pages 766--777. SIAM.

\bibitem[\protect\astroncite{Liu et~al.}{2009}]{liu2009exploratory}
Liu, X.-Y., Wu, J., and Zhou, Z.-H. (2009).
\newblock Exploratory undersampling for class-imbalance learning.
\newblock {\em Systems, Man, and Cybernetics, Part B: Cybernetics, IEEE
  Transactions on}, 39(2):539--550.

\bibitem[\protect\astroncite{Liu et~al.}{2006}]{liu2006boosting}
Liu, Y., An, A., and Huang, X. (2006).
\newblock Boosting prediction accuracy on imbalanced datasets with svm
  ensembles.
\newblock In {\em Advances in Knowledge Discovery and Data Mining}, pages
  107--118. Springer.

\bibitem[\protect\astroncite{Maciejewski and
  Stefanowski}{2011}]{maciejewski2011local}
Maciejewski, T. and Stefanowski, J. (2011).
\newblock Local neighbourhood extension of smote for mining imbalanced data.
\newblock In {\em Computational Intelligence and Data Mining (CIDM), 2011 IEEE
  Symposium on}, pages 104--111. IEEE.

\bibitem[\protect\astroncite{Maheshwari et~al.}{2011}]{maheshwari2011new}
Maheshwari, S., Agrawal, J., and Sharma, S. (2011).
\newblock A new approach for classification of highly imbalanced datasets using
  evolutionary algorithms.
\newblock {\em Intl. J. Sci. Eng. Res}, 2:1--5.

\bibitem[\protect\astroncite{Maloof}{2003}]{maloof2003learning}
Maloof, M.~A. (2003).
\newblock Learning when data sets are imbalanced and when costs are unequal and
  unknown.
\newblock In {\em ICML-2003 workshop on learning from imbalanced data sets II},
  volume~2, pages 2--1.

\bibitem[\protect\astroncite{Manevitz and Yousef}{2002}]{manevitz2002one}
Manevitz, L. and Yousef, M. (2002).
\newblock One-class svms for document classification.
\newblock {\em the Journal of machine Learning research}, 2:139--154.

\bibitem[\protect\astroncite{Mani and Zhang}{2003}]{mani2003knn}
Mani, I. and Zhang, J. (2003).
\newblock knn approach to unbalanced data distributions: a case study involving
  information extraction.
\newblock In {\em Proceedings of Workshop on Learning from Imbalanced
  Datasets}.

\bibitem[\protect\astroncite{Mart{\'\i}nez-Garc{\'\i}a
  et~al.}{2012}]{martinez2012sneom}
Mart{\'\i}nez-Garc{\'\i}a, J.~M., Su{\'a}rez-Araujo, C.~P., and B{\'a}ez, P.~G.
  (2012).
\newblock Sneom: a sanger network based extended over-sampling method.
  application to imbalanced biomedical datasets.
\newblock In {\em Neural Information Processing}, pages 584--592. Springer.

\bibitem[\protect\astroncite{Mease et~al.}{2007}]{mease2007cost}
Mease, D., Wyner, A., and Buja, A. (2007).
\newblock Cost-weighted boosting with jittering and over/under-sampling:
  Jous-boost.
\newblock {\em J. Machine Learning Research}, 8:409--439.

\bibitem[\protect\astroncite{Menardi and Torelli}{2010}]{menardi2010training}
Menardi, G. and Torelli, N. (2010).
\newblock Training and assessing classification rules with imbalanced data.
\newblock {\em Data Mining and Knowledge Discovery}, pages 1--31.

\bibitem[\protect\astroncite{Metz}{1978}]{metz1978basic}
Metz, C.~E. (1978).
\newblock Basic principles of roc analysis.
\newblock In {\em Seminars in nuclear medicine}, volume~8, pages 283--298.
  Elsevier.

\bibitem[\protect\astroncite{Mi}{2013}]{mi2013imbalanced}
Mi, Y. (2013).
\newblock Imbalanced classification based on active learning smote.
\newblock {\em Research Journal of Applied Sciences}, 5.

\bibitem[\protect\astroncite{Mladenic and
  Grobelnik}{1999}]{mladenic1999feature}
Mladenic, D. and Grobelnik, M. (1999).
\newblock Feature selection for unbalanced class distribution and naive bayes.
\newblock In {\em ICML}, volume~99, pages 258--267.

\bibitem[\protect\astroncite{Naganjaneyulu and
  Kuppa}{2013}]{naganjaneyulu2013novel}
Naganjaneyulu, S. and Kuppa, M.~R. (2013).
\newblock A novel framework for class imbalance learning using intelligent
  under-sampling.
\newblock {\em Progress in Artificial Intelligence}, 2(1):73--84.

\bibitem[\protect\astroncite{Nakamura et~al.}{2013}]{nakamura2013lvq}
Nakamura, M., Kajiwara, Y., Otsuka, A., and Kimura, H. (2013).
\newblock Lvq-smote--learning vector quantization based synthetic minority
  over--sampling technique for biomedical data.
\newblock {\em BioData mining}, 6(1):16.

\bibitem[\protect\astroncite{Napiera{\l}a et~al.}{2010}]{napierala2010learning}
Napiera{\l}a, K., Stefanowski, J., and Wilk, S. (2010).
\newblock Learning from imbalanced data in presence of noisy and borderline
  examples.
\newblock In {\em Rough Sets and Current Trends in Computing}, pages 158--167.
  Springer.

\bibitem[\protect\astroncite{Oh}{2011}]{oh2011error}
Oh, S.-H. (2011).
\newblock Error back-propagation algorithm for classification of imbalanced
  data.
\newblock {\em Neurocomputing}, 74(6):1058--1061.

\bibitem[\protect\astroncite{Pearson et~al.}{2003}]{pearson2003imbalanced}
Pearson, R., Goney, G., and Shwaber, J. (2003).
\newblock Imbalanced clustering for microarray time-series.
\newblock In {\em Proceedings of the ICML}, volume~3.

\bibitem[\protect\astroncite{Phua et~al.}{2004}]{phua2004minority}
Phua, C., Alahakoon, D., and Lee, V. (2004).
\newblock Minority report in fraud detection: classification of skewed data.
\newblock {\em ACM SIGKDD Explorations Newsletter}, 6(1):50--59.

\bibitem[\protect\astroncite{Prati et~al.}{2004a}]{prati2004class}
Prati, R.~C., Batista, G.~E., and Monard, M.~C. (2004a).
\newblock Class imbalances versus class overlapping: an analysis of a learning
  system behavior.
\newblock In {\em MICAI 2004: Advances in Artificial Intelligence}, pages
  312--321. Springer.

\bibitem[\protect\astroncite{Prati et~al.}{2004b}]{prati2004learning}
Prati, R.~C., Batista, G.~E., and Monard, M.~C. (2004b).
\newblock Learning with class skews and small disjuncts.
\newblock In {\em Advances in Artificial Intelligence--SBIA 2004}, pages
  296--306. Springer.

\bibitem[\protect\astroncite{Provost and Fawcett}{1997}]{provost1997analysis}
Provost, F.~J. and Fawcett, T. (1997).
\newblock Analysis and visualization of classifier performance: Comparison
  under imprecise class and cost distributions.
\newblock In {\em KDD}, volume~97, pages 43--48.

\bibitem[\protect\astroncite{Provost et~al.}{1998}]{PFK98}
Provost, F.~J., Fawcett, T., and Kohavi, R. (1998).
\newblock The case against accuracy estimation for comparing induction
  algorithms.
\newblock In {\em ICML'98: Proc. of the 15th Int. Conf. on Machine Learning},
  pages 445--453. Morgan Kaufmann Publishers.

\bibitem[\protect\astroncite{Ramentol et~al.}{2012a}]{ramentol2012smoteRSB}
Ramentol, E., Caballero, Y., Bello, R., and Herrera, F. (2012a).
\newblock Smote-rsb*: a hybrid preprocessing approach based on oversampling and
  undersampling for high imbalanced data-sets using smote and rough sets
  theory.
\newblock {\em Knowledge and Information Systems}, 33(2):245--265.

\bibitem[\protect\astroncite{Ramentol et~al.}{2012b}]{ramentol2012smote}
Ramentol, E., Verbiest, N., Bello, R., Caballero, Y., Cornelis, C., and
  Herrera, F. (2012b).
\newblock Smote-frst: a new resampling method using fuzzy rough set theory.
\newblock In {\em 10th International FLINS conference on uncertainty modelling
  in knowledge engineering and decision making (to appear)}.

\bibitem[\protect\astroncite{Ranawana and Palade}{2006}]{ranawana2006optimized}
Ranawana, R. and Palade, V. (2006).
\newblock Optimized precision-a new measure for classifier performance
  evaluation.
\newblock In {\em Evolutionary Computation, 2006. CEC 2006. IEEE Congress on},
  pages 2254--2261. IEEE.

\bibitem[\protect\astroncite{Raskutti and
  Kowalczyk}{2004}]{raskutti2004extreme}
Raskutti, B. and Kowalczyk, A. (2004).
\newblock Extreme re-balancing for svms: a case study.
\newblock {\em ACM Sigkdd Explorations Newsletter}, 6(1):60--69.

\bibitem[\protect\astroncite{Ribeiro}{2011}]{Rib11}
Ribeiro, R.~P. (2011).
\newblock {\em Utility-based Regression}.
\newblock PhD thesis, Dep. Computer Science, Faculty of Sciences - University
  of Porto.

\bibitem[\protect\astroncite{Ribeiro and Torgo}{2003}]{ribeiro2003predicting}
Ribeiro, R.~P. and Torgo, L. (2003).
\newblock Predicting harmful algae blooms.
\newblock In {\em Progress in Artificial Intelligence}, pages 308--312.
  Springer.

\bibitem[\protect\astroncite{Rijsbergen}{1979}]{van1979information}
Rijsbergen, C.~V. (1979).
\newblock Information retrieval. dept. of computer science, university of
  glasgow, 2nd edition.

\bibitem[\protect\astroncite{Rodr{\'\i}guez
  et~al.}{2012}]{rodriguez2012disturbing}
Rodr{\'\i}guez, J.~J., D{\'\i}ez-Pastor, J.-F., Maudes, J., and
  Garc{\'\i}a-Osorio, C. (2012).
\newblock Disturbing neighbors ensembles of trees for imbalanced data.
\newblock In {\em Machine Learning and Applications (ICMLA), 2012 11th
  International Conference on}, volume~2, pages 83--88. IEEE.

\bibitem[\protect\astroncite{Sch{\"o}lkopf
  et~al.}{2001}]{scholkopf2001estimating}
Sch{\"o}lkopf, B., Platt, J.~C., Shawe-Taylor, J., Smola, A.~J., and
  Williamson, R.~C. (2001).
\newblock Estimating the support of a high-dimensional distribution.
\newblock {\em Neural computation}, 13(7):1443--1471.

\bibitem[\protect\astroncite{Seiffert et~al.}{2011}]{seiffert2011empirical}
Seiffert, C., Khoshgoftaar, T.~M., Van~Hulse, J., and Folleco, A. (2011).
\newblock An empirical study of the classification performance of learners on
  imbalanced and noisy software quality data.
\newblock {\em Information Sciences}.

\bibitem[\protect\astroncite{Seiffert et~al.}{2010}]{seiffert2010rusboost}
Seiffert, C., Khoshgoftaar, T.~M., Van~Hulse, J., and Napolitano, A. (2010).
\newblock Rusboost: A hybrid approach to alleviating class imbalance.
\newblock {\em Systems, Man and Cybernetics, Part A: Systems and Humans, IEEE
  Transactions on}, 40(1):185--197.

\bibitem[\protect\astroncite{Sinha and May}{2004}]{sinha2004evaluating}
Sinha, A.~P. and May, J.~H. (2004).
\newblock Evaluating and tuning predictive data mining models using receiver
  operating characteristic curves.
\newblock {\em Journal of Management Information Systems}, 21(3):249--280.

\bibitem[\protect\astroncite{Song et~al.}{2009}]{song2009improved}
Song, J., Lu, X., and Wu, X. (2009).
\newblock An improved adaboost algorithm for unbalanced classification data.
\newblock In {\em Fuzzy Systems and Knowledge Discovery, 2009. FSKD'09. Sixth
  International Conference on}, volume~1, pages 109--113. IEEE.

\bibitem[\protect\astroncite{Songwattanasiri and
  Sinapiromsaran}{2010}]{songwattanasirismoute}
Songwattanasiri, P. and Sinapiromsaran, K. (2010).
\newblock Smoute: Synthetics minority over-sampling and under-sampling
  techniques for class imbalanced problem.
\newblock In {\em Proceedings of the Annual International Conference on
  Computer Science Education: Innovation and Technology, Special Track:
  Knowledge Discovery}, pages 78--83.

\bibitem[\protect\astroncite{Stefanowski and
  Wilk}{2007}]{stefanowski2007improving}
Stefanowski, J. and Wilk, S. (2007).
\newblock Improving rule based classifiers induced by modlem by selective
  pre-processing of imbalanced data.
\newblock In {\em Proc. of the RSKD Workshop at ECML/PKDD, Warsaw}, pages
  54--65.

\bibitem[\protect\astroncite{Stefanowski and
  Wilk}{2008}]{stefanowski2008selective}
Stefanowski, J. and Wilk, S. (2008).
\newblock Selective pre-processing of imbalanced data for improving
  classification performance.
\newblock In {\em Data Warehousing and Knowledge Discovery}, pages 283--292.
  Springer.

\bibitem[\protect\astroncite{Sun et~al.}{2007}]{sun2007cost}
Sun, Y., Kamel, M.~S., Wong, A.~K., and Wang, Y. (2007).
\newblock Cost-sensitive boosting for classification of imbalanced data.
\newblock {\em Pattern Recognition}, 40(12):3358--3378.

\bibitem[\protect\astroncite{Sun et~al.}{2009}]{sun2009classification}
Sun, Y., Wong, A.~K., and Kamel, M.~S. (2009).
\newblock Classification of imbalanced data: A review.
\newblock {\em International Journal of Pattern Recognition and Artificial
  Intelligence}, 23(04):687--719.

\bibitem[\protect\astroncite{Tan et~al.}{2003}]{tan2003multi}
Tan, A., Gilbert, D., and Deville, Y. (2003).
\newblock Multi-class protein fold classification using a new ensemble machine
  learning approach.

\bibitem[\protect\astroncite{Tang and Zhang}{2006}]{tang2006granular}
Tang, Y. and Zhang, Y.-Q. (2006).
\newblock Granular svm with repetitive undersampling for highly imbalanced
  protein homology prediction.
\newblock In {\em Granular Computing, 2006 IEEE International Conference on},
  pages 457--460. IEEE.

\bibitem[\protect\astroncite{Tang et~al.}{2009}]{tang2009svms}
Tang, Y., Zhang, Y.-Q., Chawla, N.~V., and Krasser, S. (2009).
\newblock Svms modeling for highly imbalanced classification.
\newblock {\em Systems, Man, and Cybernetics, Part B: Cybernetics, IEEE
  Transactions on}, 39(1):281--288.

\bibitem[\protect\astroncite{Tao et~al.}{2006}]{tao2006asymmetric}
Tao, D., Tang, X., Li, X., and Wu, X. (2006).
\newblock Asymmetric bagging and random subspace for support vector
  machines-based relevance feedback in image retrieval.
\newblock {\em Pattern Analysis and Machine Intelligence, IEEE Transactions
  on}, 28(7):1088--1099.

\bibitem[\protect\astroncite{Thai-Nghe et~al.}{2011}]{thai2011new}
Thai-Nghe, N., Gantner, Z., and Schmidt-Thieme, L. (2011).
\newblock A new evaluation measure for learning from imbalanced data.
\newblock In {\em Neural Networks (IJCNN), The 2011 International Joint
  Conference on}, pages 537--542. IEEE.

\bibitem[\protect\astroncite{Tomek}{1976}]{tomek1976two}
Tomek, I. (1976).
\newblock Two modifications of cnn.
\newblock {\em IEEE Trans. Syst. Man Cybern.}, (11):769--772.

\bibitem[\protect\astroncite{Torgo}{2005}]{T05}
Torgo, L. (2005).
\newblock Regression error characteristic surfaces.
\newblock In {\em KDD'05: Proc. of the 11th ACM SIGKDD Int. Conf. on Knowledge
  Discovery and Data Mining}, pages 697--702. ACM Press.

\bibitem[\protect\astroncite{Torgo and Ribeiro}{2003}]{torgo2003predicting}
Torgo, L. and Ribeiro, R.~P. (2003).
\newblock Predicting outliers.
\newblock In {\em Knowledge Discovery in Databases: PKDD 2003}, pages 447--458.
  Springer.

\bibitem[\protect\astroncite{Torgo and Ribeiro}{2007}]{TR07}
Torgo, L. and Ribeiro, R.~P. (2007).
\newblock Utility-based regression.
\newblock In {\em PKDD'07: Proc. of 11th European Conf. on Principles and
  Practice of Knowledge Discovery in Databases}, pages 597--604. Springer.

\bibitem[\protect\astroncite{Torgo and Ribeiro}{2009}]{TR09}
Torgo, L. and Ribeiro, R.~P. (2009).
\newblock Precision and recall in regression.
\newblock In {\em DS'09: 12th Int. Conf. on Discovery Science}, pages 332--346.
  Springer.

\bibitem[\protect\astroncite{Torgo et~al.}{2013}]{torgo2013smote}
Torgo, L., Ribeiro, R.~P., Pfahringer, B., and Branco, P. (2013).
\newblock Smote for regression.
\newblock In {\em Progress in Artificial Intelligence}, pages 378--389.
  Springer.

\bibitem[\protect\astroncite{Van Der~Putten and
  Van~Someren}{2004}]{van2004bias}
Van Der~Putten, P. and Van~Someren, M. (2004).
\newblock A bias-variance analysis of a real world learning problem: The coil
  challenge 2000.
\newblock {\em Machine Learning}, 57(1-2):177--195.

\bibitem[\protect\astroncite{Vasu and Ravi}{2011}]{vasu2011hybrid}
Vasu, M. and Ravi, V. (2011).
\newblock A hybrid under-sampling approach for mining unbalanced datasets:
  applications to banking and insurance.
\newblock {\em International Journal of Data Mining, Modelling and Management},
  3(1):75--105.

\bibitem[\protect\astroncite{Verbiest et~al.}{2012}]{verbiest2012improving}
Verbiest, N., Ramentol, E., Cornelis, C., and Herrera, F. (2012).
\newblock Improving smote with fuzzy rough prototype selection to detect noise
  in imbalanced classification data.
\newblock In {\em Advances in Artificial Intelligence--IBERAMIA 2012}, pages
  169--178. Springer.

\bibitem[\protect\astroncite{Veropoulos
  et~al.}{1999}]{veropoulos1999controlling}
Veropoulos, K., Campbell, C., and Cristianini, N. (1999).
\newblock Controlling the sensitivity of support vector machines.
\newblock In {\em Proceedings of the international joint conference on
  artificial intelligence}, volume 1999, pages 55--60. Citeseer.

\bibitem[\protect\astroncite{Wang and Japkowicz}{2010}]{wang2010boosting}
Wang, B.~X. and Japkowicz, N. (2010).
\newblock Boosting support vector machines for imbalanced data sets.
\newblock {\em Knowledge and information systems}, 25(1):1--20.

\bibitem[\protect\astroncite{Wang}{2008}]{wang2008combination}
Wang, H.-Y. (2008).
\newblock Combination approach of smote and biased-svm for imbalanced datasets.
\newblock In {\em Neural Networks, 2008. IJCNN 2008.(IEEE World Congress on
  Computational Intelligence). IEEE International Joint Conference on}, pages
  228--231. IEEE.

\bibitem[\protect\astroncite{Wang and Yao}{2009}]{wang2009diversity}
Wang, S. and Yao, X. (2009).
\newblock Diversity analysis on imbalanced data sets by using ensemble models.
\newblock In {\em Computational Intelligence and Data Mining, 2009. CIDM'09.
  IEEE Symposium on}, pages 324--331. IEEE.

\bibitem[\protect\astroncite{Wasikowski and
  Chen}{2010}]{wasikowski2010combating}
Wasikowski, M. and Chen, X.-w. (2010).
\newblock Combating the small sample class imbalance problem using feature
  selection.
\newblock {\em Knowledge and Data Engineering, IEEE Transactions on},
  22(10):1388--1400.

\bibitem[\protect\astroncite{Weiguo et~al.}{2012}]{weiguo2012improved}
Weiguo, D., Li, W., Yiyang, W., and Zhong, Q. (2012).
\newblock An improved svm-km model for imbalanced datasets.
\newblock In {\em Industrial Control and Electronics Engineering (ICICEE), 2012
  International Conference on}, pages 100--103. IEEE.

\bibitem[\protect\astroncite{Weiss}{2004}]{W04}
Weiss, G.~M. (2004).
\newblock Mining with rarity: a unifying framework.
\newblock {\em SIGKDD Explorations Newsletter}, 6(1):7--19.

\bibitem[\protect\astroncite{Weiss}{2005}]{weiss2005mining}
Weiss, G.~M. (2005).
\newblock Mining with rare cases.
\newblock In {\em Data Mining and Knowledge Discovery Handbook}, pages
  765--776. Springer.

\bibitem[\protect\astroncite{Weiss}{2010}]{weiss2010impact}
Weiss, G.~M. (2010).
\newblock The impact of small disjuncts on classifier learning.
\newblock In {\em Data Mining}, pages 193--226. Springer.

\bibitem[\protect\astroncite{Weiss and Hirsh}{2000}]{weiss2000quantitative}
Weiss, G.~M. and Hirsh, H. (2000).
\newblock A quantitative study of small disjuncts.
\newblock In {\em AAAI/IAAI}, pages 665--670.

\bibitem[\protect\astroncite{Weiss and Provost}{2003}]{weiss2003learning}
Weiss, G.~M. and Provost, F.~J. (2003).
\newblock Learning when training data are costly: the effect of class
  distribution on tree induction.
\newblock {\em J. Artif. Intell. Res.(JAIR)}, 19:315--354.

\bibitem[\protect\astroncite{Wu and Chang}{2003}]{wu2003class}
Wu, G. and Chang, E.~Y. (2003).
\newblock Class-boundary alignment for imbalanced dataset learning.
\newblock In {\em ICML 2003 workshop on learning from imbalanced data sets II,
  Washington, DC}, pages 49--56.

\bibitem[\protect\astroncite{Wu and Chang}{2005}]{wu2005kba}
Wu, G. and Chang, E.~Y. (2005).
\newblock Kba: Kernel boundary alignment considering imbalanced data
  distribution.
\newblock {\em Knowledge and Data Engineering, IEEE Transactions on},
  17(6):786--795.

\bibitem[\protect\astroncite{Xiao et~al.}{2012}]{xiao2012dynamic}
Xiao, J., Xie, L., He, C., and Jiang, X. (2012).
\newblock Dynamic classifier ensemble model for customer classification with
  imbalanced class distribution.
\newblock {\em Expert Systems with Applications}, 39(3):3668--3675.

\bibitem[\protect\astroncite{Xuan et~al.}{2013}]{xuan2013cluster}
Xuan, L., Zhigang, C., and Fan, Y. (2013).
\newblock Exploring of clustering algorithm on class-imbalanced data.
\newblock In {\em Computer Science \& Education (ICCSE), 2013 8th International
  Conference on}, pages 89--93. IEEE.

\bibitem[\protect\astroncite{Yang and Gao}{2012}]{yang2012active}
Yang, Z. and Gao, D. (2012).
\newblock An active under-sampling approach for imbalanced data classification.
\newblock In {\em Computational Intelligence and Design (ISCID), 2012 Fifth
  International Symposium on}, volume~2, pages 270--273. IEEE.

\bibitem[\protect\astroncite{Yen and Lee}{2006}]{yen2006under}
Yen, S.-J. and Lee, Y.-S. (2006).
\newblock Under-sampling approaches for improving prediction of the minority
  class in an imbalanced dataset.
\newblock In {\em Intelligent Control and Automation}, pages 731--740.
  Springer.

\bibitem[\protect\astroncite{Yen and Lee}{2009}]{yen2009cluster}
Yen, S.-J. and Lee, Y.-S. (2009).
\newblock Cluster-based under-sampling approaches for imbalanced data
  distributions.
\newblock {\em Expert Systems with Applications}, 36(3):5718--5727.

\bibitem[\protect\astroncite{Yong}{2012}]{yong2012research}
Yong, Y. (2012).
\newblock The research of imbalanced data set of sample sampling method based
  on k-means cluster and genetic algorithm.
\newblock {\em Energy Procedia}, 17:164--170.

\bibitem[\protect\astroncite{Yoon and Kwek}{2005}]{yoon2005unsupervised}
Yoon, K. and Kwek, S. (2005).
\newblock An unsupervised learning approach to resolving the data imbalanced
  issue in supervised learning problems in functional genomics.
\newblock In {\em Hybrid Intelligent Systems, 2005. HIS'05. Fifth International
  Conference on}, pages 6--pp. IEEE.

\bibitem[\protect\astroncite{Yuanhong et~al.}{2009}]{yuanhong2009cost}
Yuanhong, D., Hongchang, C., and Tao, P. (2009).
\newblock Cost-sensitive support vector machine based on weighted attribute.
\newblock In {\em Information Technology and Applications, 2009. IFITA'09.
  International Forum on}, volume~1, pages 690--692. IEEE.

\bibitem[\protect\astroncite{Zadrozny et~al.}{2003}]{zadrozny2003cost}
Zadrozny, B., Langford, J., and Abe, N. (2003).
\newblock Cost-sensitive learning by cost-proportionate example weighting.
\newblock In {\em Data Mining, 2003. ICDM 2003. Third IEEE International
  Conference on}, pages 435--442. IEEE.

\bibitem[\protect\astroncite{Zellner}{1986}]{zellner1986bayesian}
Zellner, A. (1986).
\newblock Bayesian estimation and prediction using asymmetric loss functions.
\newblock {\em Journal of the American Statistical Association},
  81(394):446--451.

\bibitem[\protect\astroncite{Zhang et~al.}{2011}]{zhang2011novel}
Zhang, D., Liu, W., Gong, X., and Jin, H. (2011).
\newblock A novel improved smote resampling algorithm based on fractal.
\newblock {\em Journal of Computational Information Systems}, 7(6):2204--2211.

\bibitem[\protect\astroncite{Zhao et~al.}{2011}]{zhao2011extended}
Zhao, H., Sinha, A.~P., and Bansal, G. (2011).
\newblock An extended tuning method for cost-sensitive regression and
  forecasting.
\newblock {\em Decision Support Systems}, 51(3):372--383.

\bibitem[\protect\astroncite{Zheng et~al.}{2004}]{zheng2004feature}
Zheng, Z., Wu, X., and Srihari, R. (2004).
\newblock Feature selection for text categorization on imbalanced data.
\newblock {\em ACM SIGKDD Explorations Newsletter}, 6(1):80--89.

\bibitem[\protect\astroncite{Zhou and Liu}{2006}]{zhou2006training}
Zhou, Z.-H. and Liu, X.-Y. (2006).
\newblock Training cost-sensitive neural networks with methods addressing the
  class imbalance problem.
\newblock {\em Knowledge and Data Engineering, IEEE Transactions on},
  18(1):63--77.

\bibitem[\protect\astroncite{Zhu and Hovy}{2007}]{zhu2007active}
Zhu, J. and Hovy, E.~H. (2007).
\newblock Active learning for word sense disambiguation with methods for
  addressing the class imbalance problem.
\newblock In {\em EMNLP-CoNLL}, volume~7, pages 783--790.

\bibitem[\protect\astroncite{Zhuang and Dai}{2006a}]{zhuang2006parameterestim}
Zhuang, L. and Dai, H. (2006a).
\newblock Parameter estimation of one-class svm on imbalance text
  classification.
\newblock In {\em Advances in Artificial Intelligence}, pages 538--549.
  Springer.

\bibitem[\protect\astroncite{Zhuang and Dai}{2006b}]{zhuang2006parameter}
Zhuang, L. and Dai, H. (2006b).
\newblock Parameter optimization of kernel-based one-class classifier on
  imbalance learning.
\newblock {\em Journal of Computers}, 1(7):32--40.

\end{thebibliography}






\end{document}